\documentclass[jkps,nofootinbib,twocolumn,showkeys,superscriptaddress]{revtex4-2}
\usepackage[pdftex]{graphicx}
\usepackage{amssymb}
\usepackage{amsmath}
\usepackage{txfonts}
\usepackage{bm}

\begin{document}
\setcounter{page}{1}
\title[]{Effects of relational graph modularity and depth on the learning performance of neural networks}
\author{Yash \surname{Arya}}
\author{Sang Hoon \surname{Lee}}
\email{lshlj82@gnu.ac.kr}
\thanks{Fax: +82-55-772-1409}
\affiliation{Department of Physics, Gyeongsang National University, Jinju, 52828, Korea}
\date[]{Received \today}

\begin{abstract}
In recent years, graph-based machine learning techniques, such as reinforcement learning and graph neural networks, have garnered significant attention. While some recent studies have started to explore the relationship between the graph structure of neural networks and their predictive performance, they often limit themselves to a narrow range of model networks, particularly lacking mesoscale structures such as communities. Our work advances this area by conducting a more comprehensive investigation, incorporating realistic network structures characterized by heterogeneous degree distributions and community structures, which are typical characteristics of many real networks. These community structures offer a nuanced perspective on network architecture. Our analysis employs model networks such as random and scale-free networks, alongside a comparison with a biological neural network and its subsets for more detailed analysis. We examine the impact of these structural attributes on the performance of image classification tasks. Our findings reveal that structural properties do affect performance to some extent. Specifically, within moderate-depth architectures, networks featuring coherent, densely interconnected communities demonstrate enhanced learning capabilities. Crucially, we find that this advantage is strictly depth-dependent: extending the architecture to eight layers reverses the effect entirely. This comparison with the biological neural network emphasizes the relevance of our findings to real-world structures, suggesting an intriguing connection worth further exploration. This study contributes meaningfully to network science and machine learning, providing insights that could inspire the design of more biologically informed neural networks.
\end{abstract}



\maketitle

\section{Introduction}
\label{sec:intro}
The predictive accuracy of a neural network is fundamentally influenced by its underlying graph structure, where neurons are represented as nodes connected by edges that facilitate information transmission through multiple hidden layers~\cite{LeCun2015DeepLearning}. Despite considerable advancements and breakthroughs over the past decade, understanding the relationship between neural network performance and its graph structure remains one of the most challenging questions in machine learning today~\cite{you2020design,zoph2017neural}. One way to tackle this problem is by focusing on real-world networks, particularly biological neural networks and harnessing their efficient information processing~\cite{dressler2010survey} and graph structures to improve the design of artificial neural networks.

Real-world networks, ranging from social and technological systems to biological neural circuits, universally exhibit a common set of structural properties: heterogeneous degree distributions consistent with power-law degree distribution~\cite{barabasi1999emergence}, small-world topology characterized by high clustering and short average path lengths~\cite{watts1998collective}, and mesoscale community organization where nodes with similar attributes connect more densely within modular subgraphs~\cite{Porter2009,Fortunato_review,Fortunato2022}. Biological neural networks are among the most thoroughly documented examples of systems exhibiting all three properties simultaneously~\cite{bullmore2009complex,meunier2010modular}. These findings suggest that the structural properties of biological neural networks, namely sparsity, communities, and heterogeneous degree distributions co-occur and thus may contribute to high-performance information processing in the most rigorously optimized natural systems known~\cite{clune2013evolutionary}.

This raises a natural and important scientific question: do these structural properties offer computational advantages that are specific to biological systems, or are they general enough to improve information processing in artificial neural networks as well? 

A notable approach called \texttt{graph2nn}~\cite{jiaxuan2020} introduced a relational graph representation to answer this question by modeling neural networks as message exchange functions on graphs. This enables the exploration of various neural architectures, such as multilayer perceptrons (MLPs)~\cite{MLP}, convolutional neural networks (CNNs)~\cite{CNN}, and residual neural networks (ResNets)~\cite{ResNet}, demonstrating improved predictive performance on data sets such as Canadian Institute For Advanced Research (CIFAR-10)~\cite{cifar10} and ImageNet~\cite{ImageNet}. The study identifies key relational graph characteristics, such as clustering coefficient and average path length, that contribute to enhanced performance.

Our work takes a crucial step forward by addressing unresolved questions and extending the methodology to incorporate a broader spectrum of network structures~\cite{BarabasiBook,NewmanBook,FirstCourseBook}. We emphasize the importance of graph structures at the layer level of neural networks, integrating more heterogeneous and realistic network configurations. These configurations are characterized by different degree distributions, such as those found in ER random networks~\cite{erdos1959random} and scale-free networks generated by the static model~\cite{goh2001}, which are emblematic of many real-world networks. Additionally, we enhance these structures by imposing community structures~\cite{Porter2009,Fortunato_review,Fortunato2022}, which are mesoscale arrangements prevalent in numerous real networks, including brain networks. Our findings reveal that structural properties do affect performance to some extent. 

Specifically, networks featuring coherent, densely interconnected communities demonstrate enhanced learning capabilities. This comprehensive expansion not only deepens the understanding of neural network architecture but also bridges the gap between theoretical models and practical, real-world applications, providing new insights into optimizing neural network performance. Please note that we do not assume that realistic structural properties improve neural network performance, rather we test whether they do. The fully-connected network, which imposes no structural prior and represents the simplest possible architecture, serves as the baseline against which all performance comparisons are made throughout this study. This distinction between investigating a hypothesis and assuming its conclusion is central to the scientific approach taken here. Our findings hence indicate the possibility that incorporating these diverse and realistic network structures into neural network design can provide insights into optimizing performance, thereby contributing to the broader understanding of the interplay between network topology and machine learning outcomes.

The paper is organized as follows. In Sec.~\ref{sec:method}, we introduce our main method for incorporating network structures into relational graphs used in learning processes. We present the type of network structures we use in Sec.~\ref{sec:networks}, followed by our results in Sec.~\ref{sec:results}. Finally, we conclude the paper in Sec.~\ref{sec:conclusion}.

\begin{figure*}
    \begin{tabular}{l}
    (a) \\
    \includegraphics[trim=1.6cm 4.3cm 1cm 4cm, clip=true, width=\textwidth]{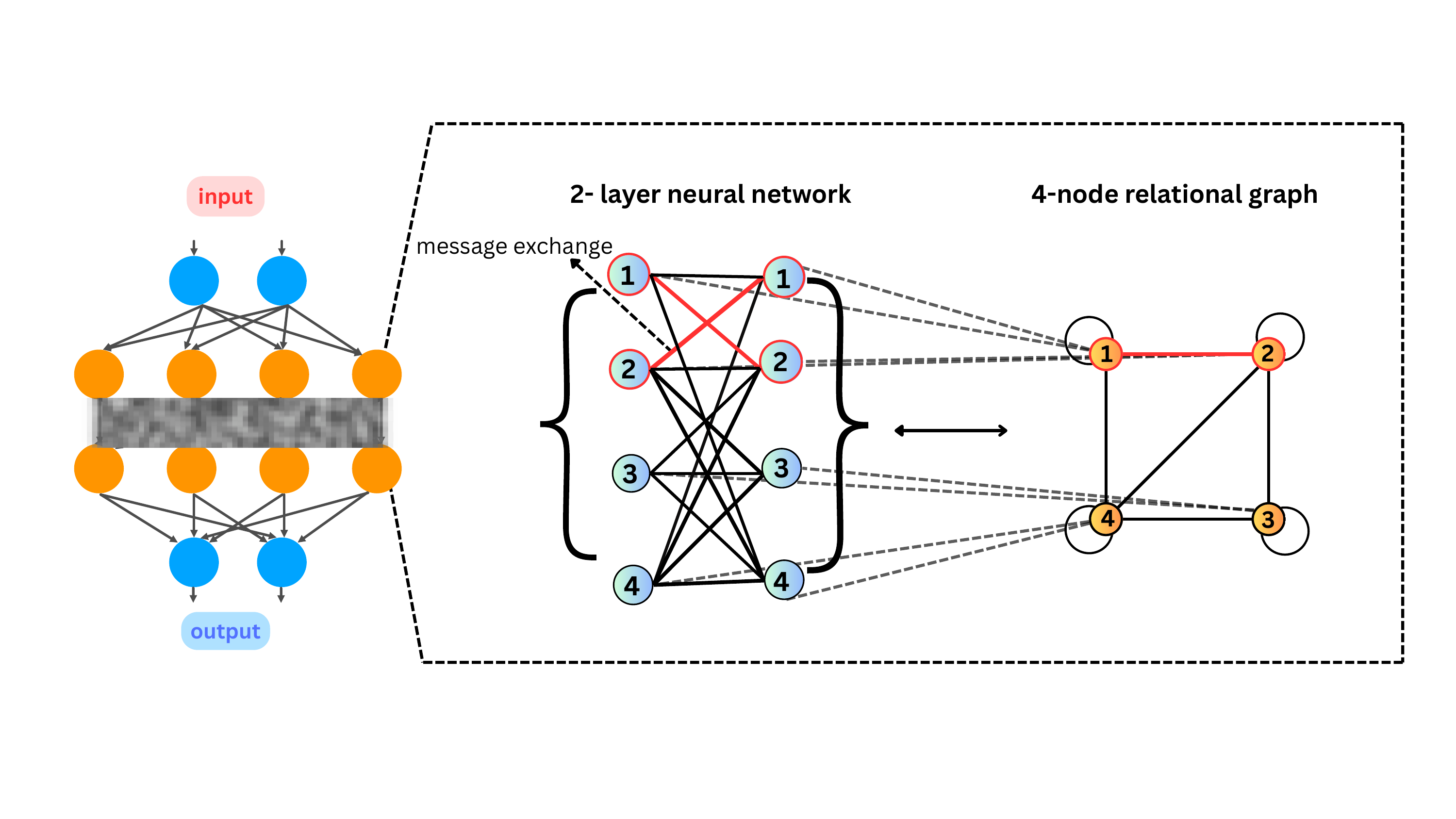} \\
    (b) \\
    \includegraphics[trim=0.5cm 4cm 2cm 4cm, clip=true, width=\textwidth]{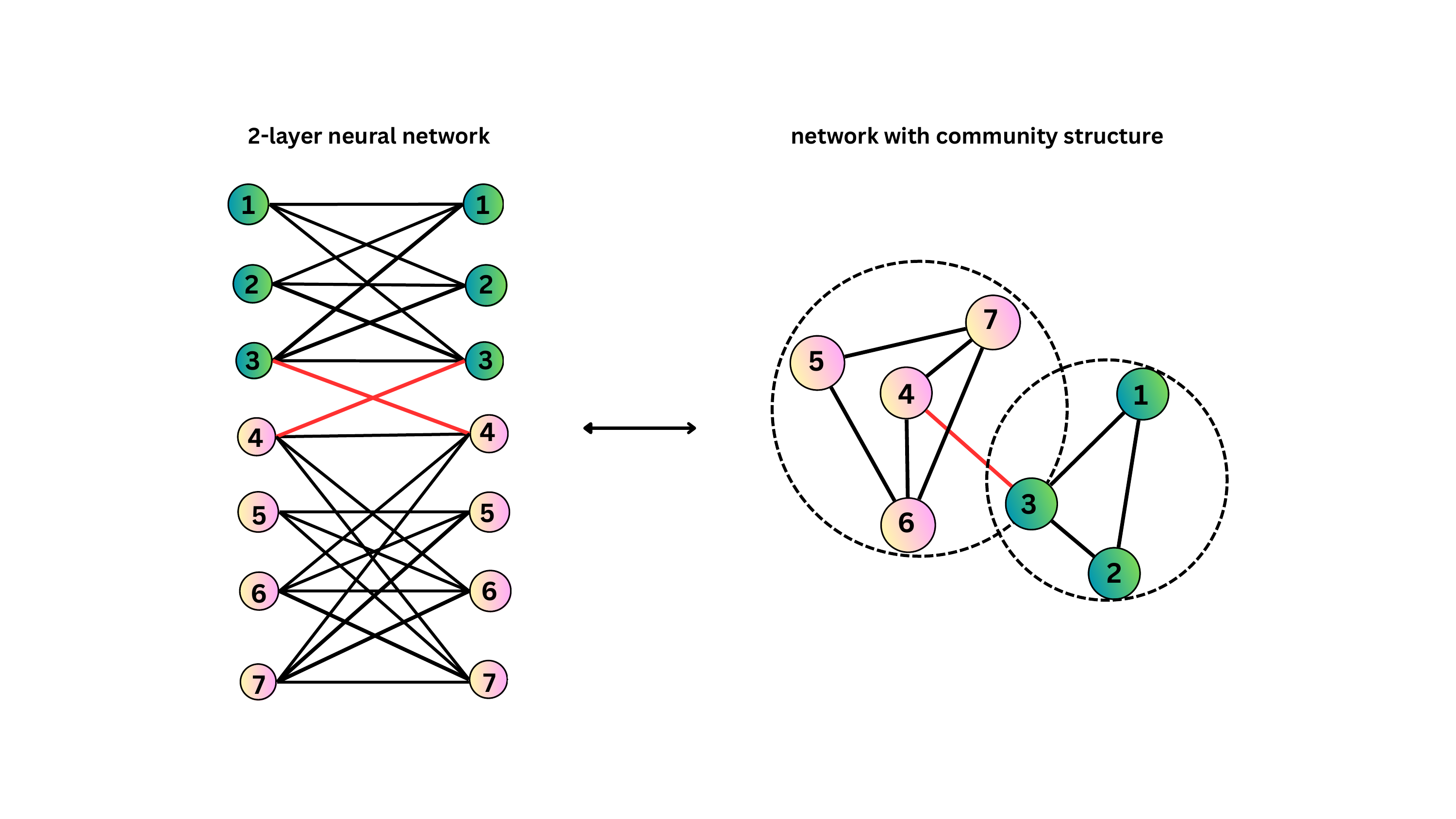}
    \end{tabular}
    \caption{Schematic illustration of the translation between multi-layer perceptron (MLP) architectures and relational graphs. (a)~The upper panel demonstrates the mapping of a 2-layer neural network block into a 4-node relational graph via dimensional reduction based on message exchange. In this example, only the edge between nodes 1 and 3 is absent, while all other connections---including self-loops---are present by default. A self-loop on a node in the relational graph corresponds to a direct inter-layer connection between the identical neuron index across successive layers in the MLP representation. For clarity, the specific connection between node 1 and node 2 is highlighted with red edges in both frameworks. To construct deep architectures with an adjustable total depth (number of hidden layers), this baseline 2-layer computational block is sequentially repeated (stacked) for $R$ message-passing rounds, while the node dimensions of the input and output layers remain fixed by the data set dimensionality and the number of classification target classes, respectively. (b)~The lower panel illustrates the conversion of a modular relational graph featuring community structures (distinguished by light pink and green node groupings, with self-loops omitted for visual clarity) on the right into its corresponding 2-layer neural network translation on the left. The mesoscale division between community $A$ (nodes 1--3) and community $B$ (nodes 4--7) maps directly onto the MLP bipartite wiring pattern, where the sole inter-community message exchange is mediated via the edge connecting nodes 3 and 4.}
    \label{fig:schematic}
\end{figure*}

\section{The relational graph method}
\label{sec:method}
We begin by introducing the relational graph method~\cite{jiaxuan2020}, which offers a novel perspective on neural network architectures by focusing on message exchanges between neurons instead of merely directed data flow. According to this framework, a fixed-width fully-connected layer can be represented as a relational graph by representing one input channel and one output channel together as a single node, and an edge as message exchange between the two nodes. Thus, relational graphs can represent many types of neural network layers which provides a significant advantage over traditional static, directed unidirectional computational graphs. A graph $G$ is termed a relational graph when it models neural networks as dynamic systems of communication, with computations translated into rounds of message exchanges within the graph. Following the basic setting from Ref.~\cite{jiaxuan2020}, each round of message exchange corresponds to a layer in the neural network. We utilize message functions $\sigma(\cdot)$ to transform input node features into output messages along each edge. The process for the $r$-th round of message exchange is described using a vector-based formulation with the rectified linear unit (ReLU) activation function as follows:
\begin{equation}
x_i^{(r+1)} = \sigma \left( \sum_{j \in N(i)} w_{ij}^{(r)} x_j^{(r)} \right)
\end{equation}
where $x_i^{(r)}$ is the $i$-th element of the vector $\mathbf{x}^{(r)}$ of node features for node $i$ and $w_{ij}^{(r)}$ is the $(i,j)$ element of the matrix $\mathbf{W}^{(r)}$ at round $r$, and $\sigma(\cdot)$ is the ReLU activation function. The neighborhood $N(i)$ includes the node itself to account for self-edges, which is a necessary adaptation in neural-network-based machine learning, without which node $i$'s own state at message exchange round $r$ would get entirely discarded and play no role in determining its state at round $r+1$. The weight matrix $\mathbf{W}^{(r)}$ is equivalent to the weighted adjacency matrix~\cite{BarabasiBook,NewmanBook,FirstCourseBook} except for the self-edge part, and is subject to tuning during the learning process to optimize the network's performance. This tuning process involves adjusting the weights to minimize a loss function and improve the model's predictive accuracy. Since weights are not shared across rounds, each of the $R$ rounds has its own independent weight matrix $\mathbf{W}^{(r)}$, with the same sparsity pattern but distinct learned values.

This formulation is completely general with respect to graph topology, which means in contrast to traditional neural networks which are configured with fully connected layers, both \texttt{graph2nn}~\cite{jiaxuan2020} and our method utilize not-fully-connected (sparser) networks, enabling more efficient and potentially more powerful representations by incorporating diverse and realistic relational graphs. This approach extends the MLP's architecture beyond simple fully-connected configurations to include complex relational graphs that incorporate community structures and varying node features, which is crucial for modeling realistic and heterogeneous network configurations. In Ref.~\cite{jiaxuan2020}, the ``WS-flex'' graph generator, a variant of the Watts-Strogatz model from 1998~\cite{watts1998collective}, was employed to explore neural network architectures. This model is characterized by its large clustering coefficient and small average path length, defining the ``small-world'' property. However, it is limited to these characteristics and does not account for more complex network features. In contrast, our work leverages more realistic network structures, incorporating heterogeneous degree distributions (``scale-free'' networks), and community structures that more accurately reflect real-world networks~\cite{BarabasiBook,NewmanBook,FirstCourseBook}. These enhancements allow for a more comprehensive exploration of how network structure influences learning performance.

To illustrate how the relational graph method works, Fig.~\ref{fig:schematic} provides a visual representation. Figure~\ref{fig:schematic}(a) demonstrates the translation of a 2-layer neural network into a 4-node relational graph using dimensional reduction based on message exchange. In Fig.~\ref{fig:schematic}(b), the process of translating artificial networks with community structures into a relational graph using edge indices is depicted. This relational graph is then converted into a neural network with multiple rounds of layers, each corresponding to a round of message exchange. Basically, this repeated structure constitutes the \emph{hidden layers} [the masked region in the leftmost panel of Fig.~\ref{fig:schematic}(a)] of neural networks for learning~\cite{MLP,rumelhart1986,CNN,mehta2019}, and the numbers of nodes in the input and output layers correspond to the dimensions of input data and the number of output classifications, respectively [the input layer and the first hidden layer are fully connected, and so are the last hidden layer and the output layer, as depicted in Fig.~\ref{fig:schematic}(a)]. 

To empirically evaluate this framework, we deploy our relational graph configurations on the standard CIFAR-10 image classification task~\cite{cifar10}. This choice fixes the input layer at $32 \times 32 \times 3 = 3072$ nodes (corresponding to the $32 \times 32$ pixel dimensions across the red, green, and blue color channels) and the output layer at $10$ nodes (representing the target classification categories). Within these data set boundaries, we map the translated graph topologies onto a baseline fixed-width 5-layer MLP. This controlled environment enables us to systematically isolate and analyze variations in statistical learning performance as a function of the underlying network structure and its macroscale graph topological parameters. To provide further clarity on these structural mappings, we have developed an interactive web visualization~\cite{YA_relational_graph} that dynamically demonstrates the translation of neural architectures, including those with community structures, into their relational graph counterparts.

\section{Networks used for learning}
\label{sec:networks}

\subsection{Model networks}
\label{sec:real_networks}

We begin by introducing the network models used in our study. The first is the Erd\H{o}s-R{\'e}nyi (ER) random network~\cite{erdos1959random}, which is characterized by its simplicity and randomness, where each edge is included with a constant probability independently of every other edge. This model serves as a baseline for understanding network dynamics without inherent structure and is characterized by a binomial degree distribution, which follows a Poisson distribution in the sparse limit. The ER network is defined by a single parameter $p$, the connection probability, for a given number of nodes (the ``size'' of the network).

In contrast, the second model is the static scale-free network (SFN)~\cite{goh2001}, which allows us to explore networks with a power-law degree distribution $p(k) \sim k^{-\gamma}$, where the degree exponent $\gamma$ can be in a wide range $\gamma \in [2, \infty)$. This provides flexibility in controlling the degree distribution. Specifically, as $\gamma$ increases, the degree distribution becomes more homogeneous, resembling the characteristics of the ER model, whereas smaller values of $\gamma$ result in more heterogeneous networks. For $2 < \gamma \leq 3$, the mean degree is finite but the variance diverges, representing the regime of strongest heterogeneity. For $\gamma > 3$, both the mean and variance are finite. As $\gamma \to \infty$, the degree distribution converges toward a Poisson distribution, recovering a similar behavior to ER. Our experiments span $\gamma \in [2, 6]$, covering all three regimes. The static model is characterized by two parameters: $\gamma$, the degree exponent, and $m$, the average degree of the network, for a given number of nodes.

Next, we incorporate community structures~\cite{Porter2009,Fortunato_review,Fortunato2022} into these network models. Real-world networks often divide into modules, or communities, where nodes connect more densely with each other than with the rest of the network. This modularity can enhance network performance and learning efficiency by leveraging the intrinsic organization of nodes. The previous work, \texttt{graph2nn}~\cite{jiaxuan2020}, has made significant strides in exploring network architectures but lacks the integration of more sophisticated and realistic mesoscale structures such as communities. We address this by utilizing the Lancichinetti-Fortunato-Radicchi (LFR) benchmark~\cite{lancichinetti2009community} to generate synthetic networks with known community structures, allowing for heterogeneity in node degrees and community sizes.

To integrate model networks with communities, we developed a simplified version of the LFR algorithm, which is capable of generating networks with desired community structures without complex dependencies (code available at Ref.~\cite{YA_LFR}). In this framework, the global network is partitioned into a set of $c$ distinct communities, where each individual community functions as a locally isolated subgraph (modeled either as an ER random network or a static SFN). These modules are then interconnected via a mixing parameter $\mu \in [0,1]$, which is the fraction of a node's total degree allocated to edges linking to nodes in external communities, i.e., $\mu = k_i^{\text{inter}} / k_i$.

Under this planted partition, the empirical graph structure directly maps onto the classical Newman--Girvan modularity $Q$~\cite{NewmanBook}. Given $L$ total edges and $c$ statistically symmetric, equal-sized communities where each module accounts for an equal share of the total degree $K_s / (2L) = 1/c$, the fraction of intra-community edges is $\sum_{s=1}^c L_s / L = 1 - \mu$. The expected modularity therefore evaluates analytically to
\begin{equation}
Q = \sum_{s=1}^c \left[ \frac{L_s}{L} - \left(\frac{K_s}{2L}\right)^2 \right] \approx (1 - \mu) - \frac{1}{c} \,,
\end{equation}
which one can easily derive by considering each term of the summand separately\footnote{Specifically, because each node places a fraction $1 - \mu$ of its edges internally within its module, the total internal degree endpoints in each module satisfy $2L_s = (1 - \mu) K_s$, yielding $\sum_{s=1}^c (L_s / L) = (1 - \mu) \sum_{s=1}^c [K_s / (2L)] = 1 - \mu$. For equal-sized communities with identical expected degree distributions, each module accounts for an equal share of the total degree endpoints, $K_s / (2L) = 1/c$, so the configuration-model penalty evaluates to $\sum_{s=1}^c (K_s / 2L)^2 = \sum_{s=1}^c (1/c)^2 = 1/c$.}.
When $\mu \to 0$, edges are confined strictly within individual modules and $Q$ attains its theoretical maximum of $1 - 1/c$, whereas as $\mu \to 1 - 1/c$, the connectivity approaches a randomized null model without community preference ($Q \to 0$). Tuning this mixing parameter $\mu$ thus allows for the systematic generation of diverse graph configurations with precisely controlled structural modularity. We note that the aggregate modularity $Q$ itself is fundamentally unsuitable for direct cross-network comparisons across varying community numbers $c$ or system sizes, as it suffers from well-known resolution limits and intrinsic scale dependence. Consequently, the mixing parameter $\mu$ and the number of communities $c$ serve as far more fundamental, unambiguous structural control parameters than the $Q$ value.

Ultimately, the structural frameworks explored here---built upon ER random network and static SFN baselines---form a systematic, hierarchical decomposition of structural contributions rather than isolated, independent analyses. In this design, ER networks serve as the fundamental null model to establish baseline sparsity, while static scale-free architectures introduce controlled degree heterogeneity. Modular community structure is then integrated not as a separate network model, but as an independent mesoscale organizational factor superimposed onto both underlying configurations. Taken together, this incremental layering of topological properties allows us to cleanly isolate and rank the relative contributions of connectivity sparsity, degree distribution profile, and modular community organization on resulting neural network performance.

\subsection{Biological neural network}
\label{sec:biological_networks}

To deepen our understanding of the relationship between underlying graph structures and predictive performance, we incorporate the well-known biological neural network of \emph{C. elegans}, a type of nematode or roundworm~\cite{brenner1974genetics}. This network is one of the most thoroughly studied biological neural systems and provides a valuable benchmark due to its complete mapping of neurons and synapses. Notably, the \emph{C. elegans} neural network consists of 277 neurons, which is of a similar order of magnitude to the model networks we use in our study, making it particularly suitable for comparison. Our approach systematically evaluates the predictive performance of this biological neural network, contrasting with the previous study~\cite{jiaxuan2020}, where comparisons were made with a coarse-grained version of the macaque brain primarily through visual inspection of adjacency matrices compared with a small-world network model. By focusing on the \emph{C. elegans} neural network, we aim to provide a more detailed and systematic assessment.

For our analysis, we consider the entire \emph{C. elegans} brain network and the frontal neural network, as described in~\cite{kaiser2006nonoptimal}. To ensure a fair comparison, we construct an ensemble of random subnetworks by drawing $N_{\text{subgraphs}} = 5$ independent random samples of 131 nodes from the whole brain network, matching the size of its frontal cortex, and evaluate each across independent random initialization seeds. We compare their predictive performance with both the frontal neural network and ER random networks of similar sizes. These comparisons are detailed in Sec.~\ref{sec:results_C_elegans}.

\subsection{Machine-learning implementation}
\label{sec:implementation}

For implementation, we translate network configurations into a fixed-width 5-layer MLP using the method described in Sec.~\ref{sec:method}. This translation allows us to study the impact of network structure on learning performance. We conduct simulations on networks, each initially consisting of $128$ nodes, divided into $1$, $2$, \dots, $8$ uniform-sized communities (thus each community has $128$, $64$, \dots, $16$ nodes, respectively)\footnote{In practice, the actual number of nodes in each community can be smaller than the desired numbers, as we only consider the largest component in the percolation sense for each community. This approach ensures the connectivity of the entire network, preventing the isolation of message exchanges and facilitating effective information flow across the network.}. Each community within these networks is generated using the ER random network model~\cite{erdos1959random} or the static scale-free network model~\cite{goh2001}, as introduced in Sec.~\ref{sec:real_networks}. The nodes belonging to different communities are connected according to the mixing parameter $\mu$, also discussed in Sec.~\ref{sec:real_networks}. Additionally, we include simulations on the \emph{C. elegans} neural network, as described in Sec.~\ref{sec:biological_networks}. While we acknowledge that the sizes of the networks used in this study are relatively small compared to those typically analyzed in conventional network studies, this choice is necessitated by the practical challenges of computational cost, as similarly noted in previous works~\cite{xie2019exploring,jiaxuan2020}.

We perform all simulations using the CIFAR-10 image classification data set~\cite{cifar10}. Each model is trained for $200$ epochs using five different random seeds on the NVIDIA GeForce RTX 4090 graphics processing unit (GPU), equipped with CUDA Version 12.2 and PyTorch 2.5.1. We report the average results obtained from these runs. Additionally, to address the need for statistical validation within realistic computational boundaries, we selectively expand our ensemble size to ten independent random initializations (seeds) for a representative subset of parameter configurations across our phase spaces. Crucially, our cross-verification tests on these representative configurations demonstrate that the qualitative performance conclusions, structural boundaries, and trajectories of the architectural sweet spots remain completely invariant under the larger ensemble size, confirming that our mapped phase-space gradients reflect genuine topological drivers rather than stochastic sampling noise. 

By understanding the effects of integrating community structures and exploring a range of network configurations, we aim to uncover insights into the interplay between network topology and neural network learning performance. We present the results in the following section.

\section{Results}
\label{sec:results}

\begin{figure}
\includegraphics[trim=5cm 4cm 1cm 5cm, clip=true, width=0.4\textwidth]{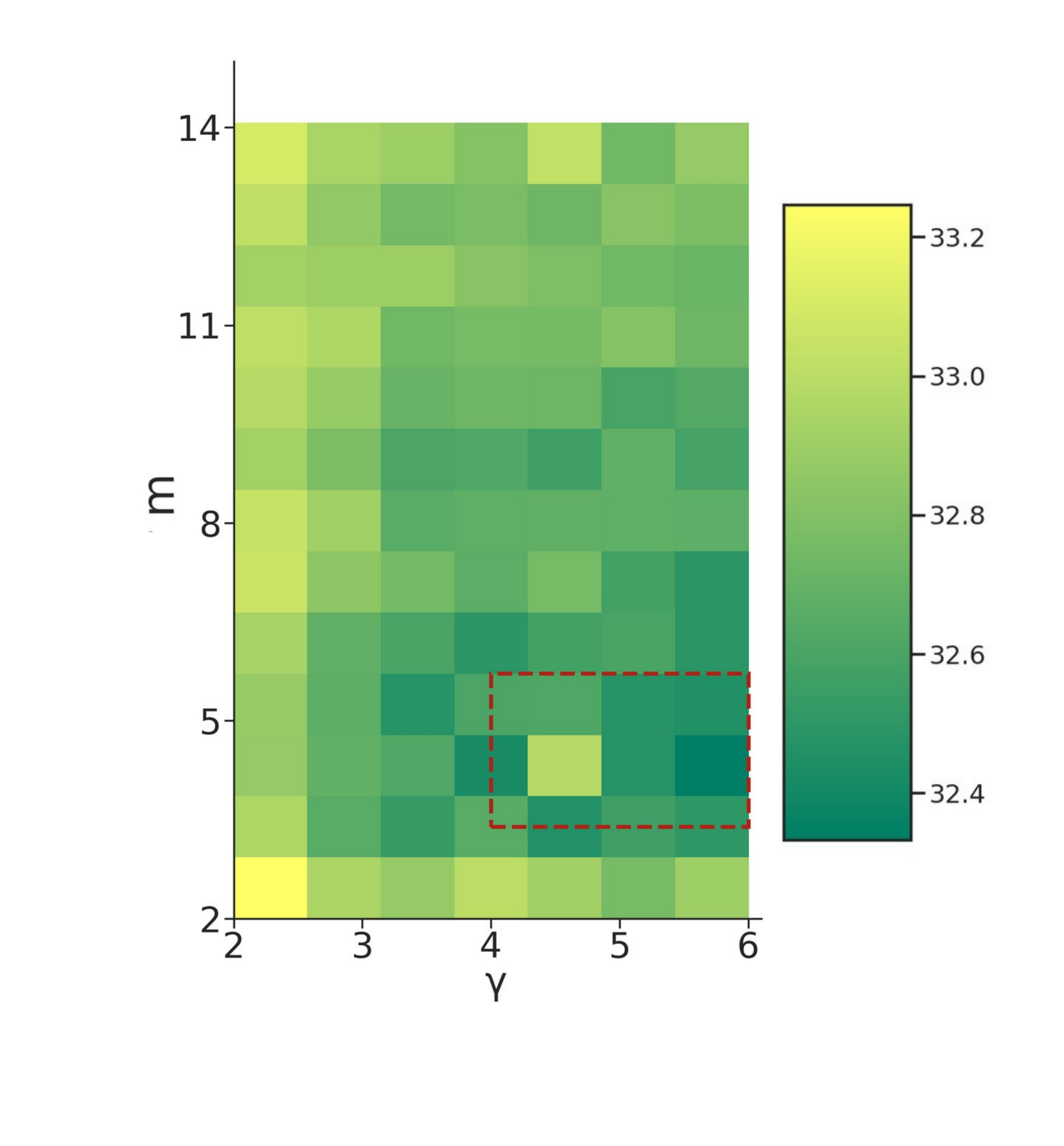}
\caption{Performance summary of static scale-free network configurations using a fixed-width 5-layer MLP on the CIFAR-10 data set. The region of optimal performance is highlighted by a red rectangular box, with the color bar indicating the top-1 error percentage (we mark the value for the baseline result as the red horizontal line). The performance for all configurations is always better than the complete-graph baseline of top-1 error $\approx 33.28\%$ (the largest value of top-1 error throughout the configurations is $\approx 33.25\%$).}
\label{fig:SFN_single}
\end{figure}

\begin{figure*}
\begin{tabular}{llll}
(a) & (b) & (c) & (d) \\
\includegraphics[width=0.23\textwidth,height=4cm]{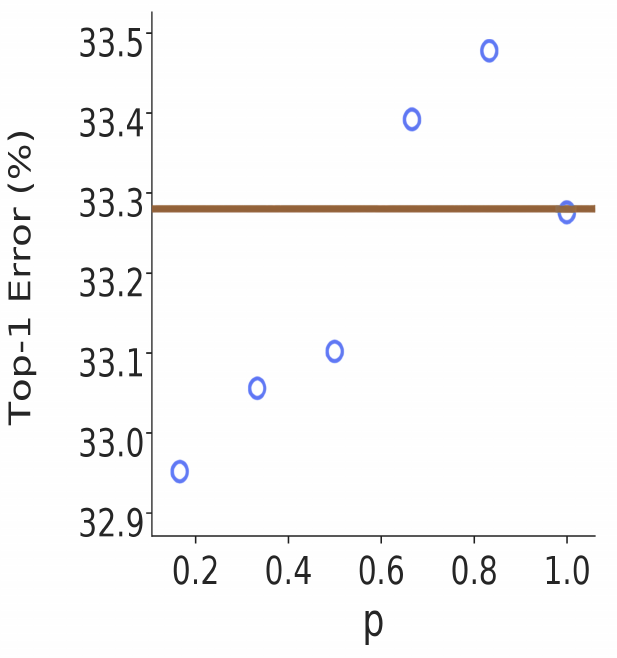} &
\includegraphics[width=0.25\textwidth,height=5cm]{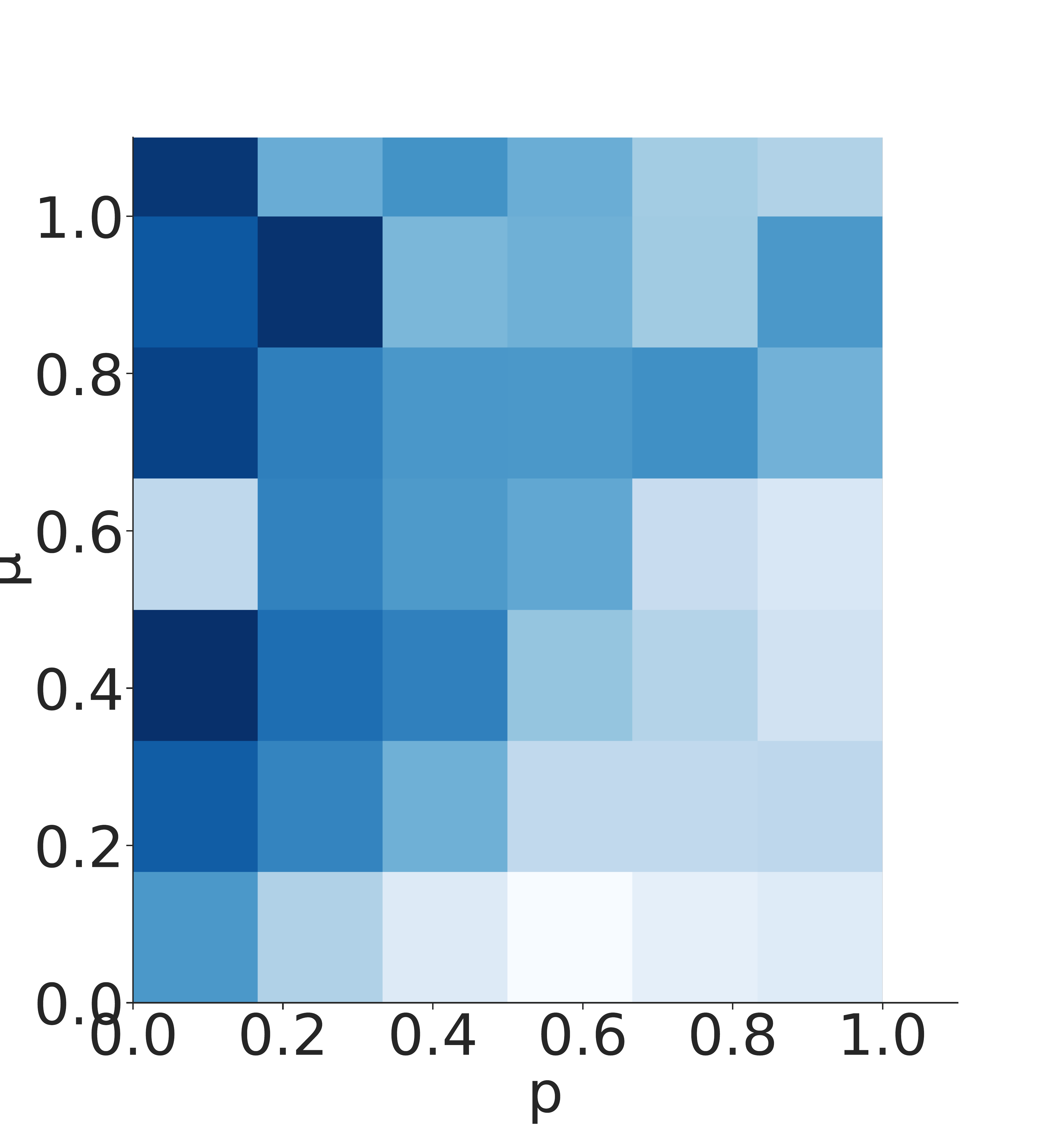} &
\includegraphics[width=0.25\textwidth,height=5cm]{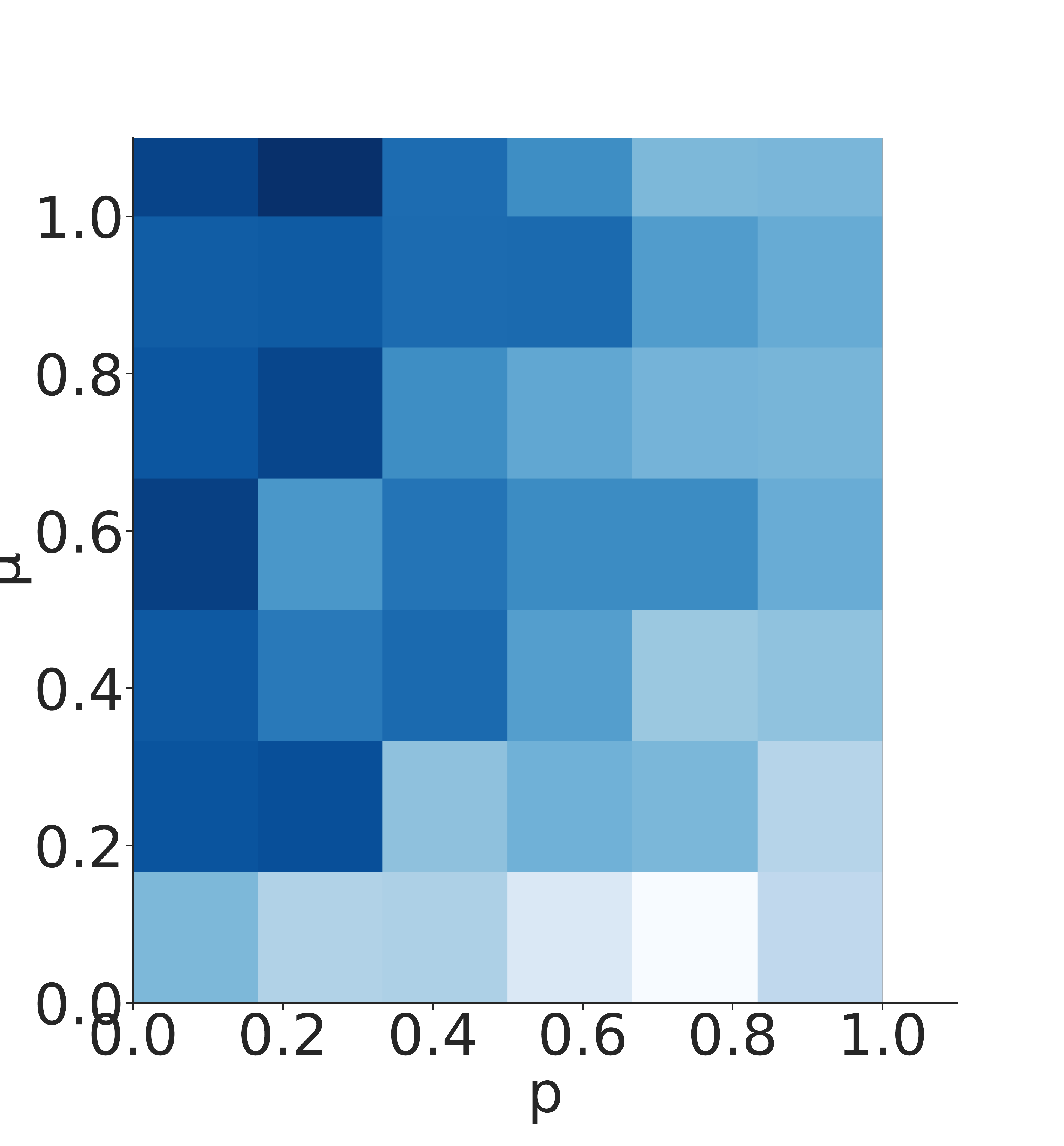} &
\includegraphics[width=0.25\textwidth,height=5cm]{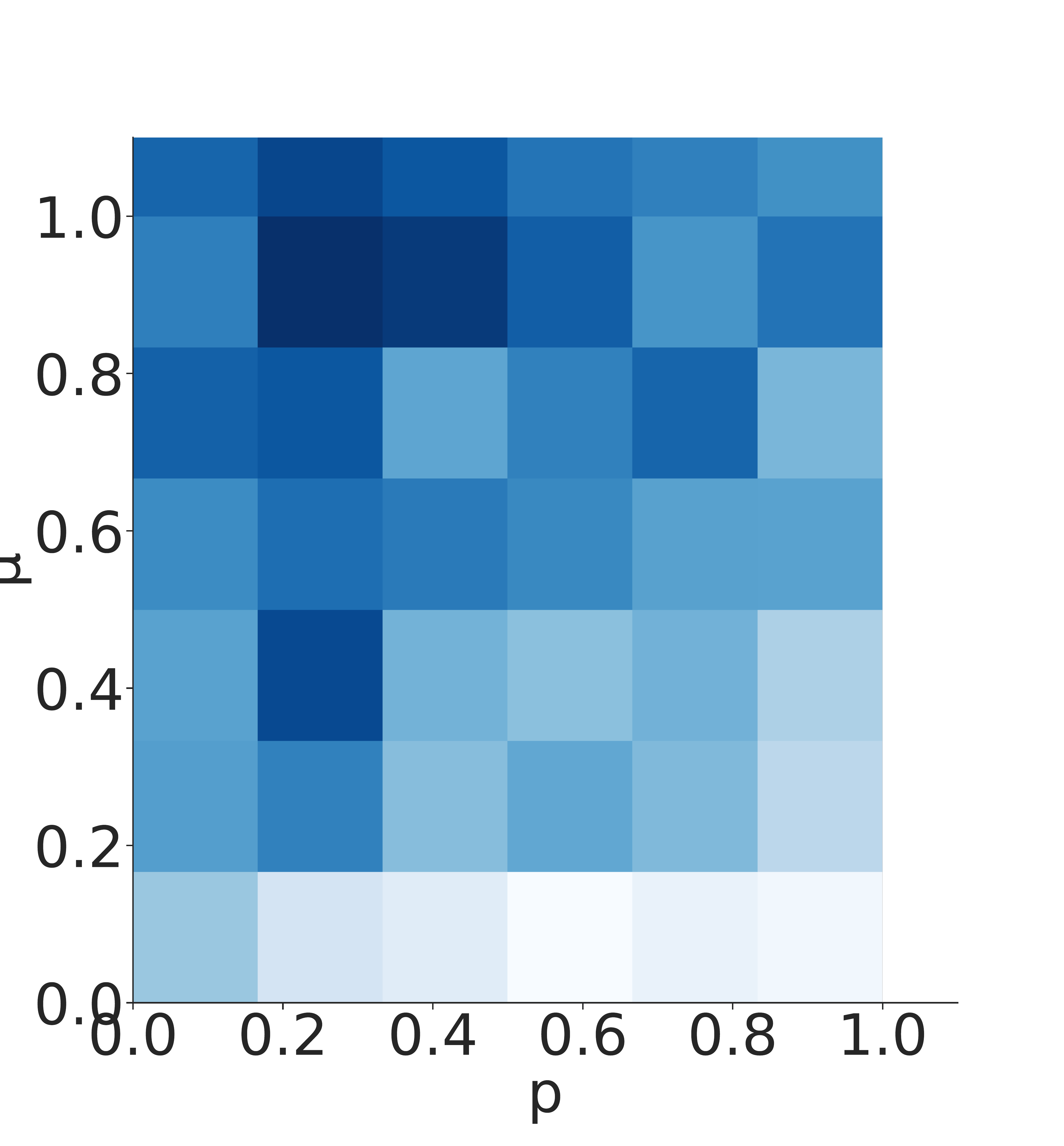} \\
(e) & (f) & (g) & (h) \\
\includegraphics[width=0.25\textwidth,height=5cm]{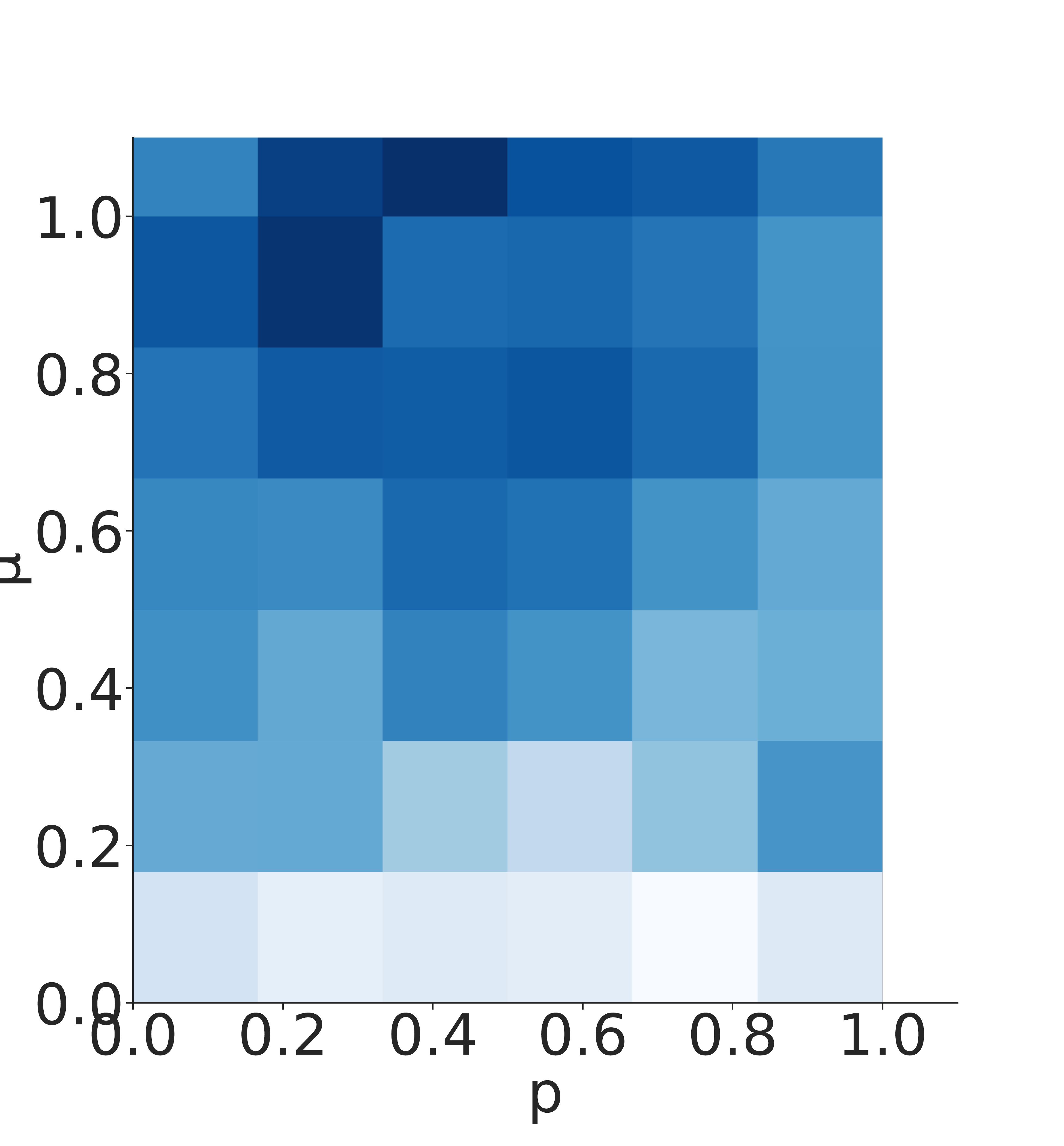} &
\includegraphics[width=0.25\textwidth,height=5cm]{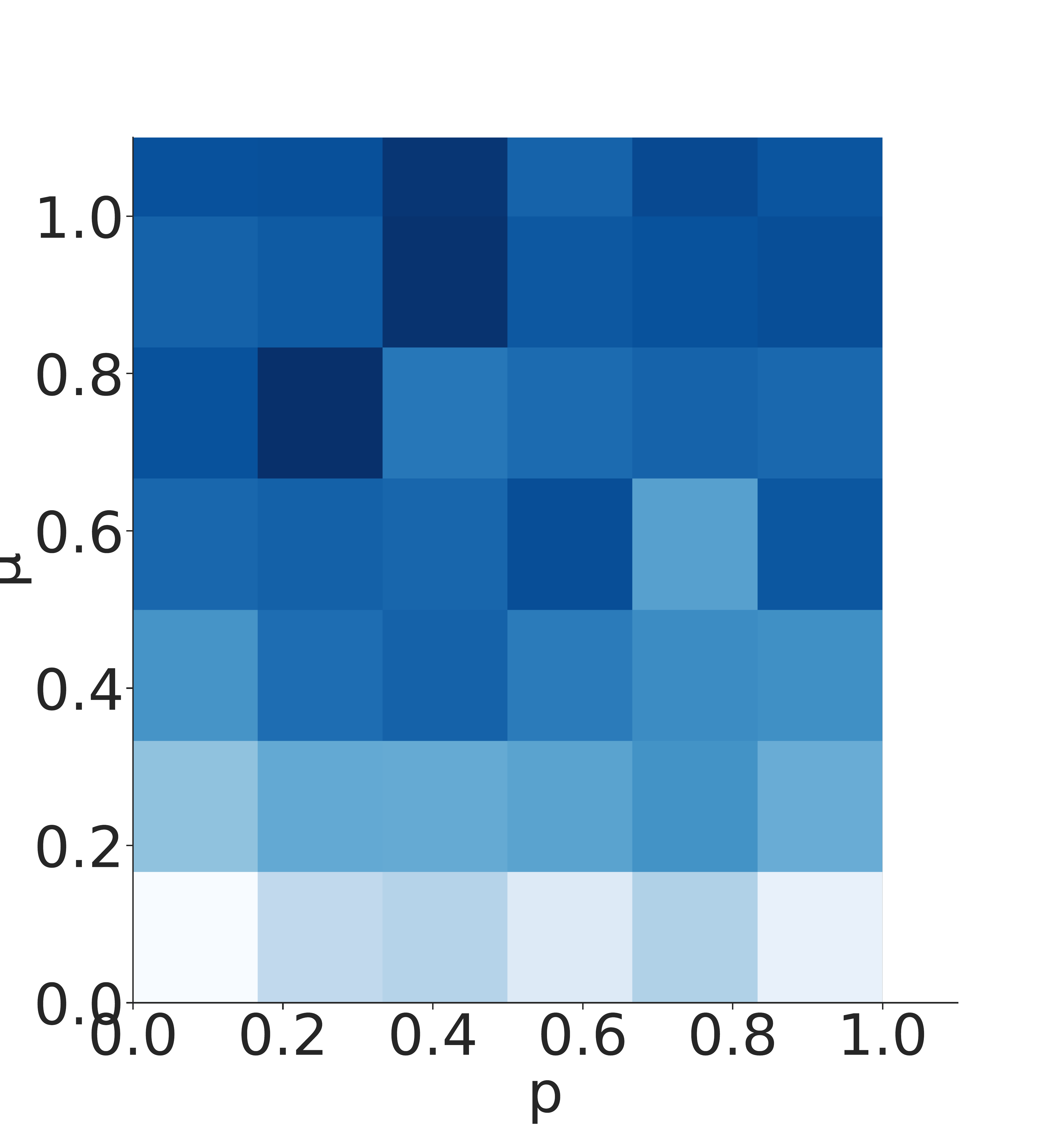} &
\includegraphics[width=0.25\textwidth,height=5cm]{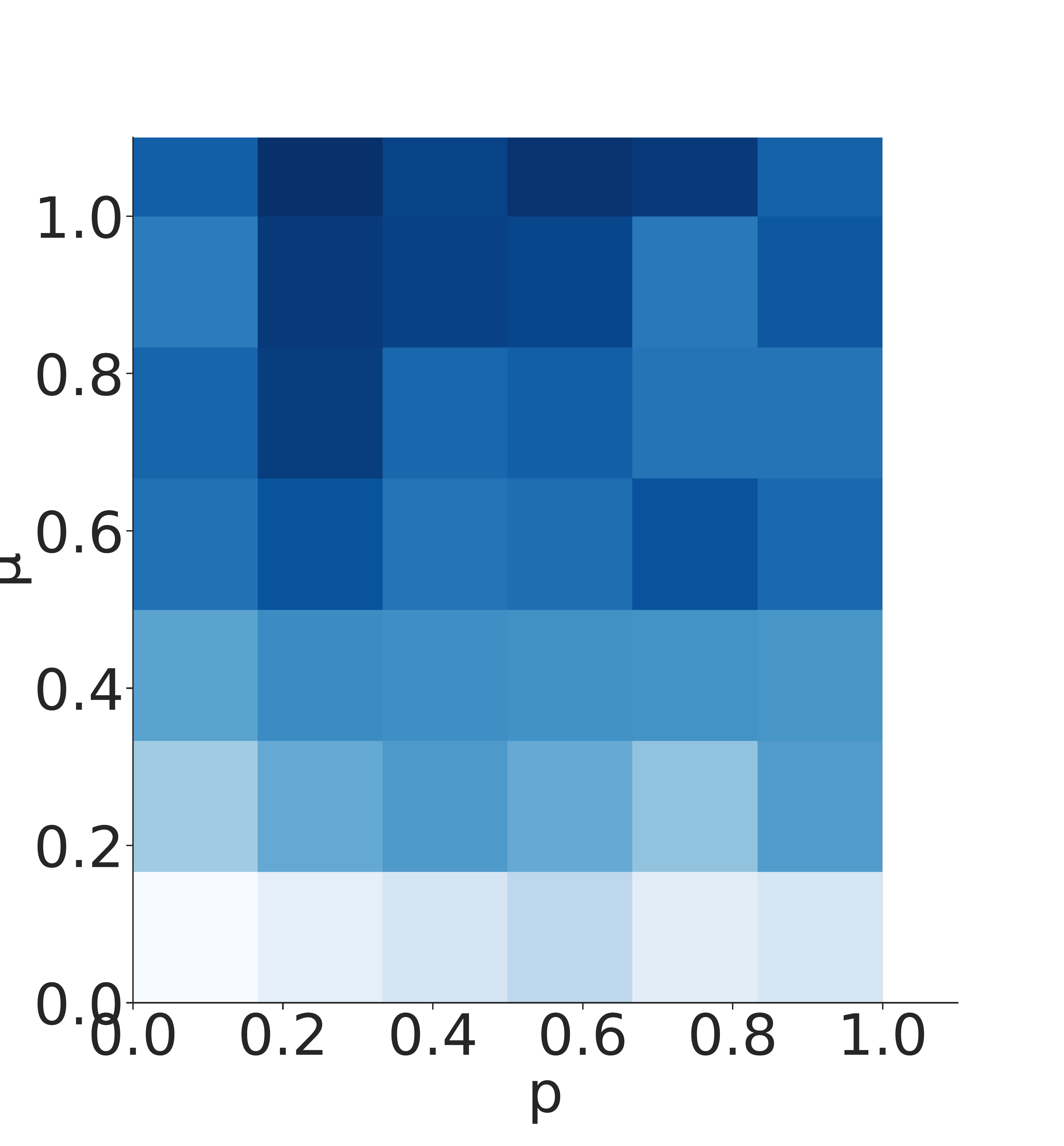} &
\includegraphics[width=0.25\textwidth,height=5cm]{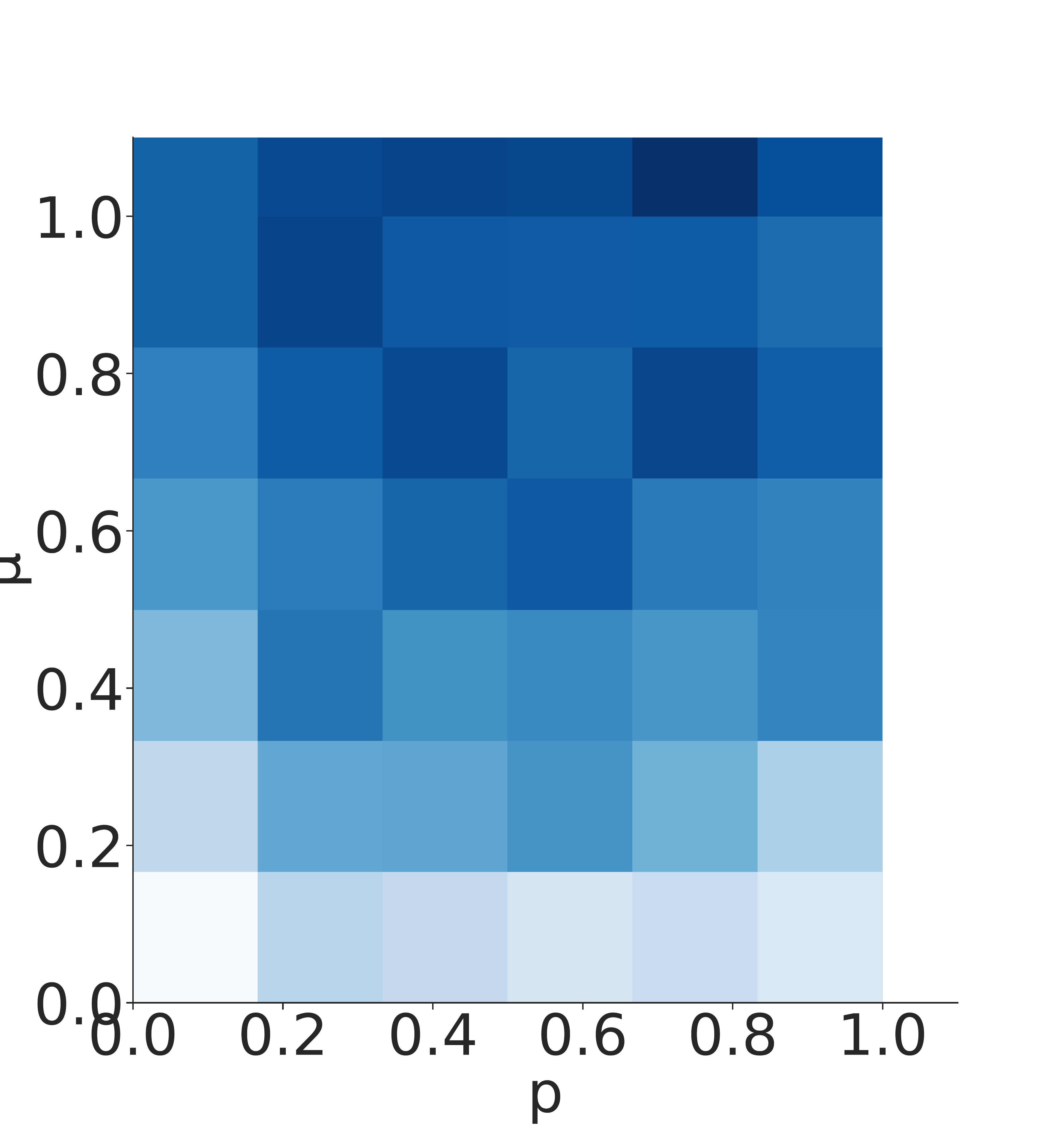} \\
\end{tabular}
\includegraphics[trim=19cm 12.2cm 23cm 13cm, clip=true, width=0.3\linewidth,height=2.2cm]{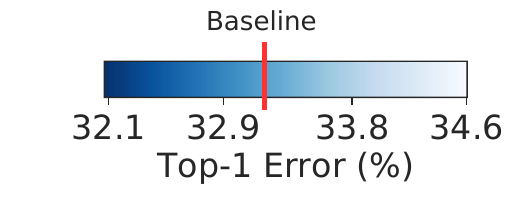}
\caption{Performance of ER random network configurations with $128$ nodes and varying numbers of communities using the fixed-width 5-layer MLP on the CIFAR-10 data set~\cite{cifar10} (a) A single community (the fully-connected network serves as the baseline shown by the brown horizontal line), (b) 2 communities, (c) 3 communities, (d) 4 communities, (e) 5 communities, (f) 6 communities, (g) 7 communities, (h) 8 communities. In the cases of multiple communities (b)--(h), the parameters for the phase diagrams are the internal connection probability $p$~\cite{erdos1959random} and the mixing parameter $\mu$~\cite{lancichinetti2009community}, and the horizontal color bar with the baseline marked indicates the performance for those cases.}
\label{fig:random_results}
\end{figure*}

\begin{figure*}
\begin{tabular}{llll}
(a) & (b) & (c) & (d) \\
\includegraphics[width=0.23\textwidth,height=4cm]{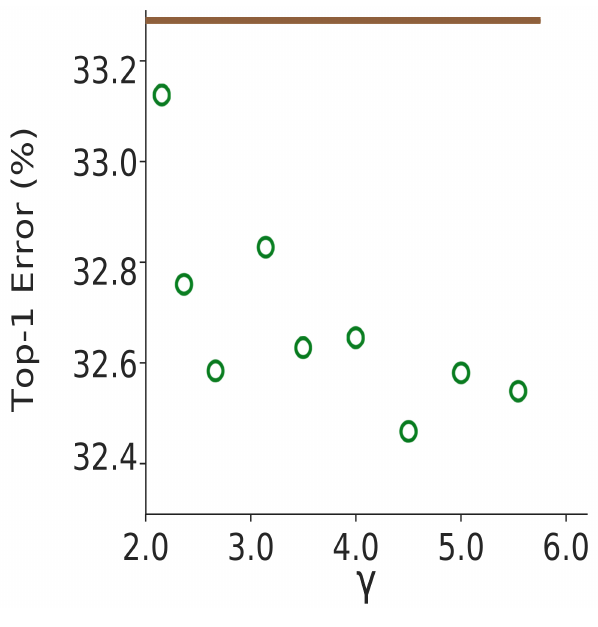} &
\includegraphics[width=0.25\textwidth,height=5cm]{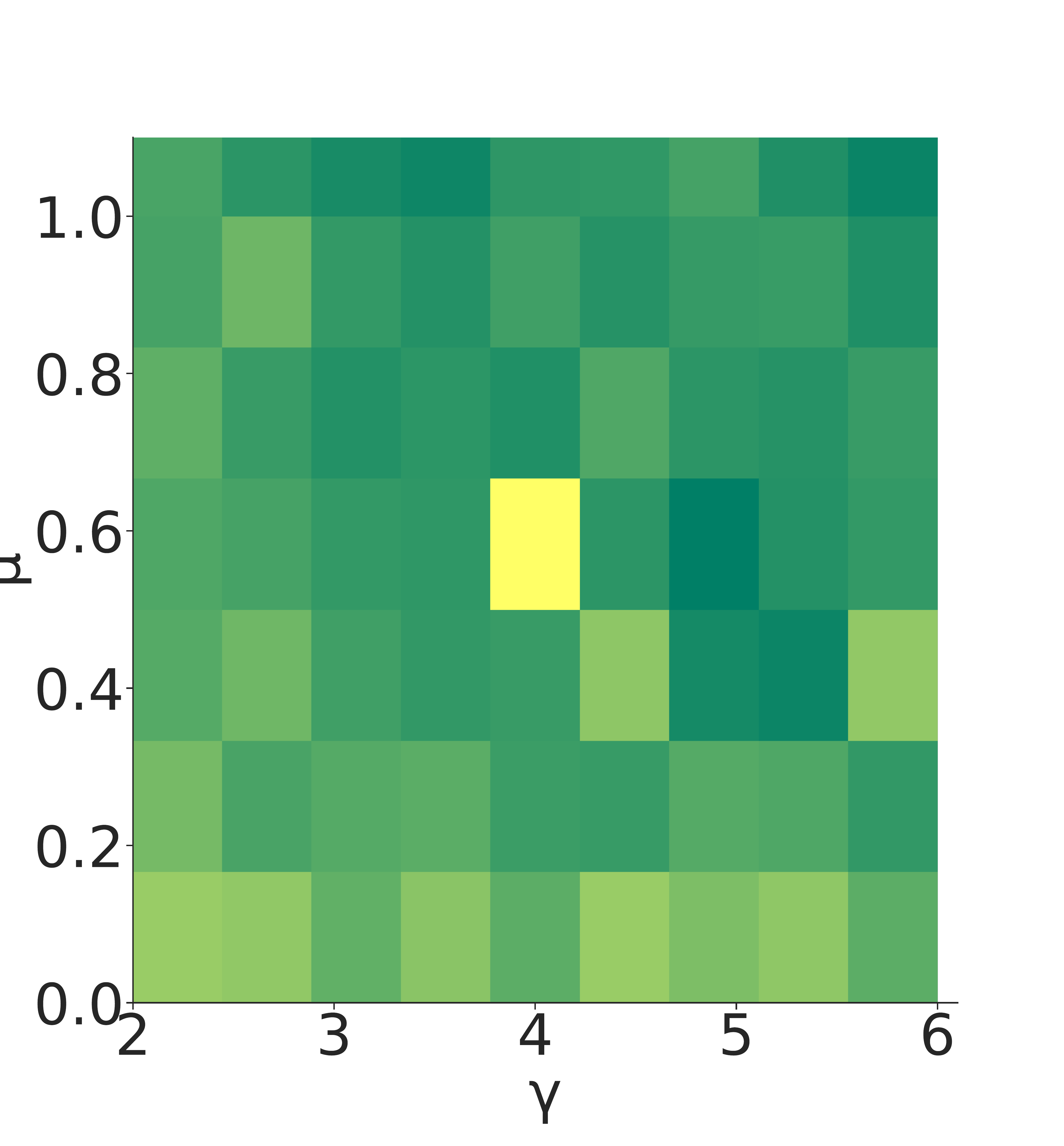} &
\includegraphics[width=0.25\textwidth,height=5cm]{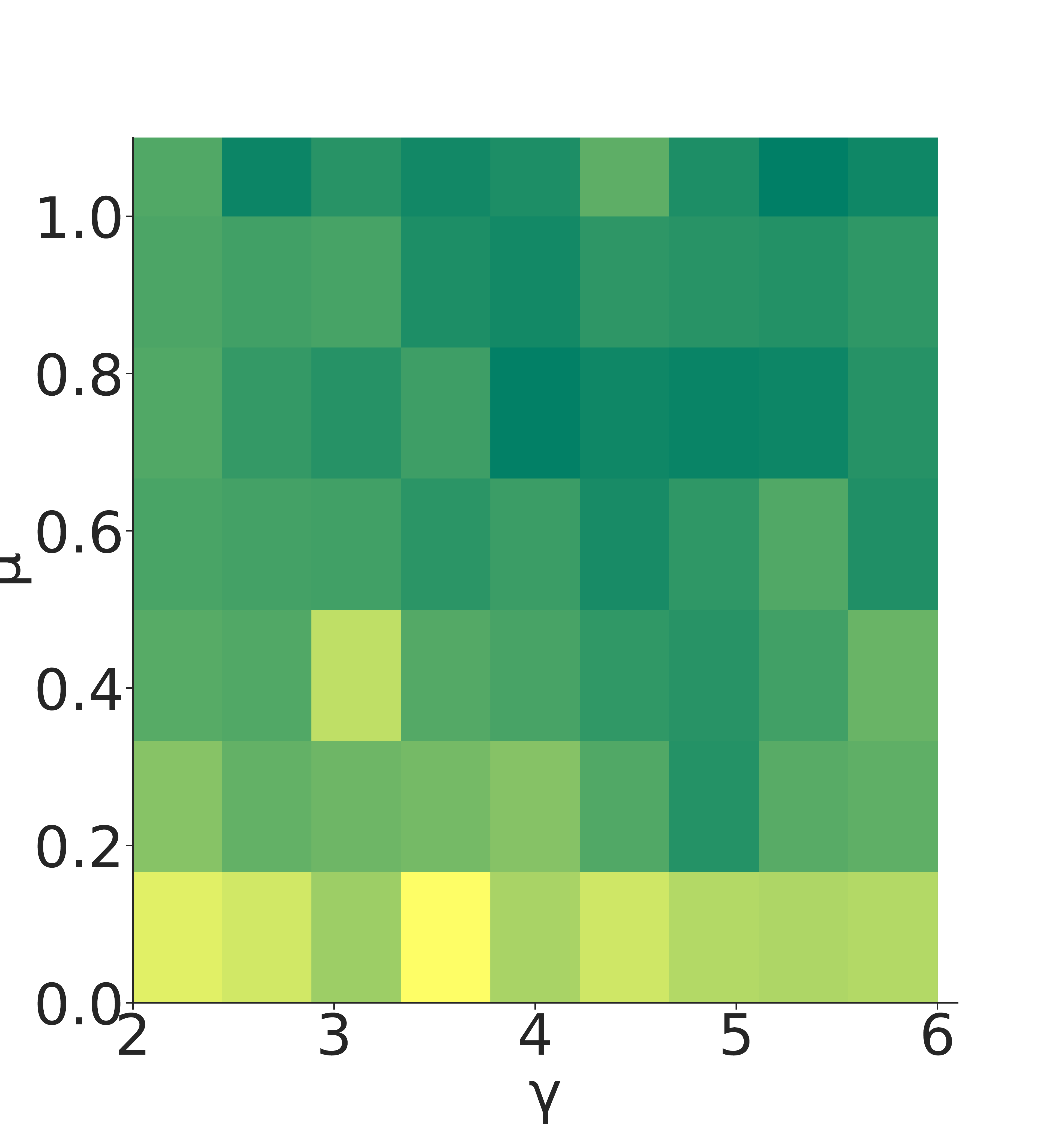} &
\includegraphics[width=0.25\textwidth,height=5cm]{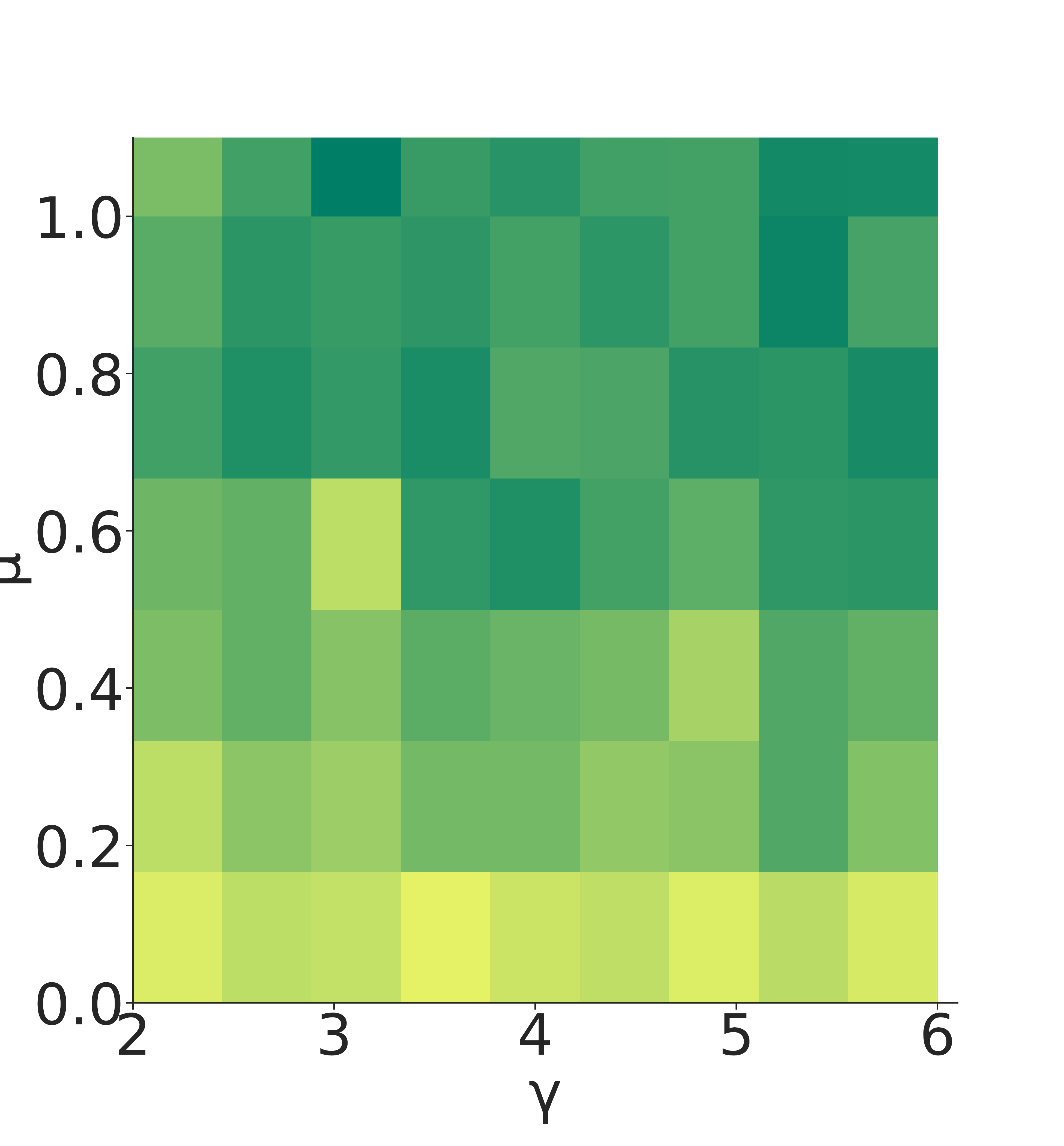} \\
(e) & (f) & (g) & (h) \\
\includegraphics[width=0.25\textwidth,height=5cm]{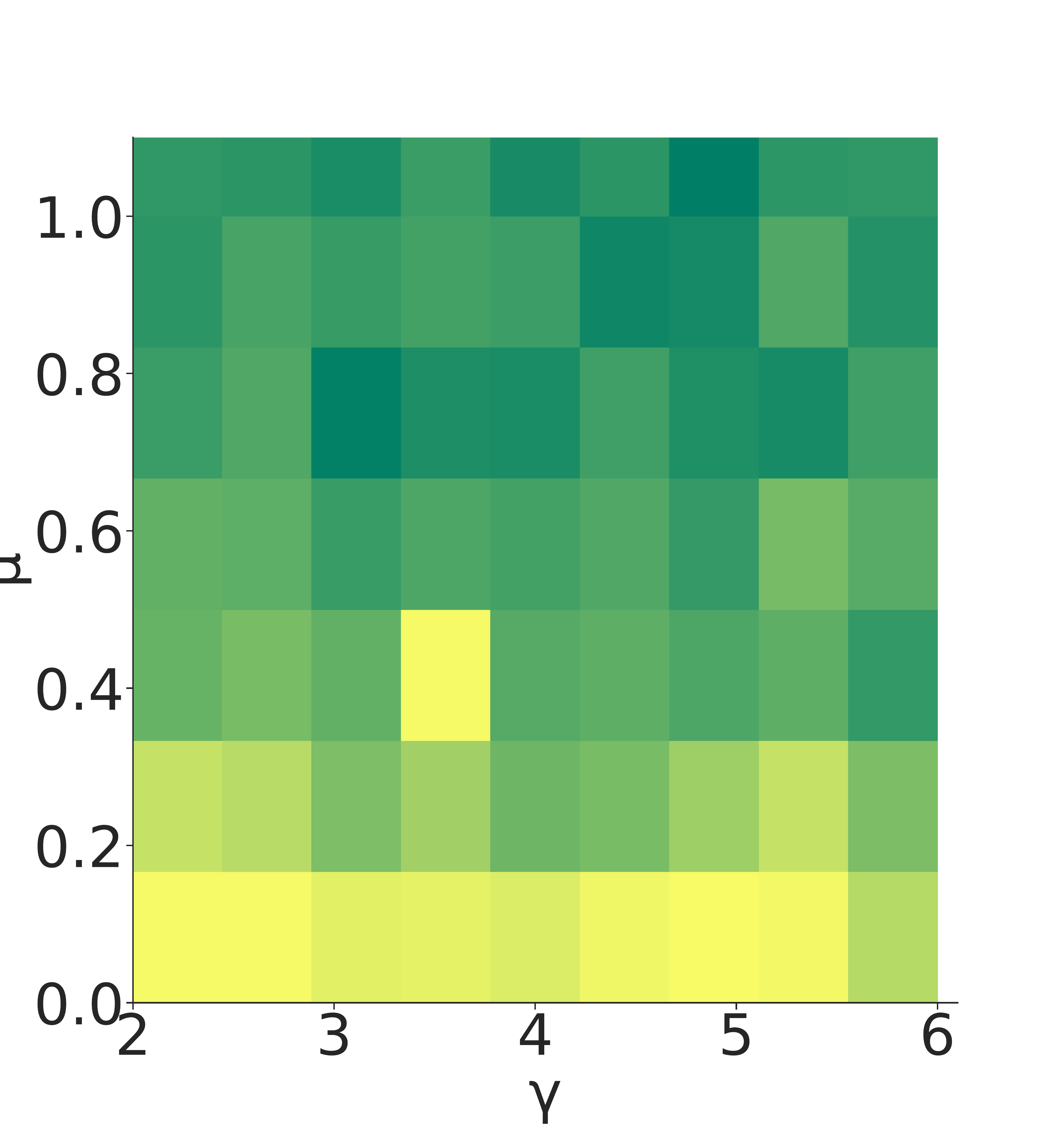} &
\includegraphics[width=0.25\textwidth,height=5cm]{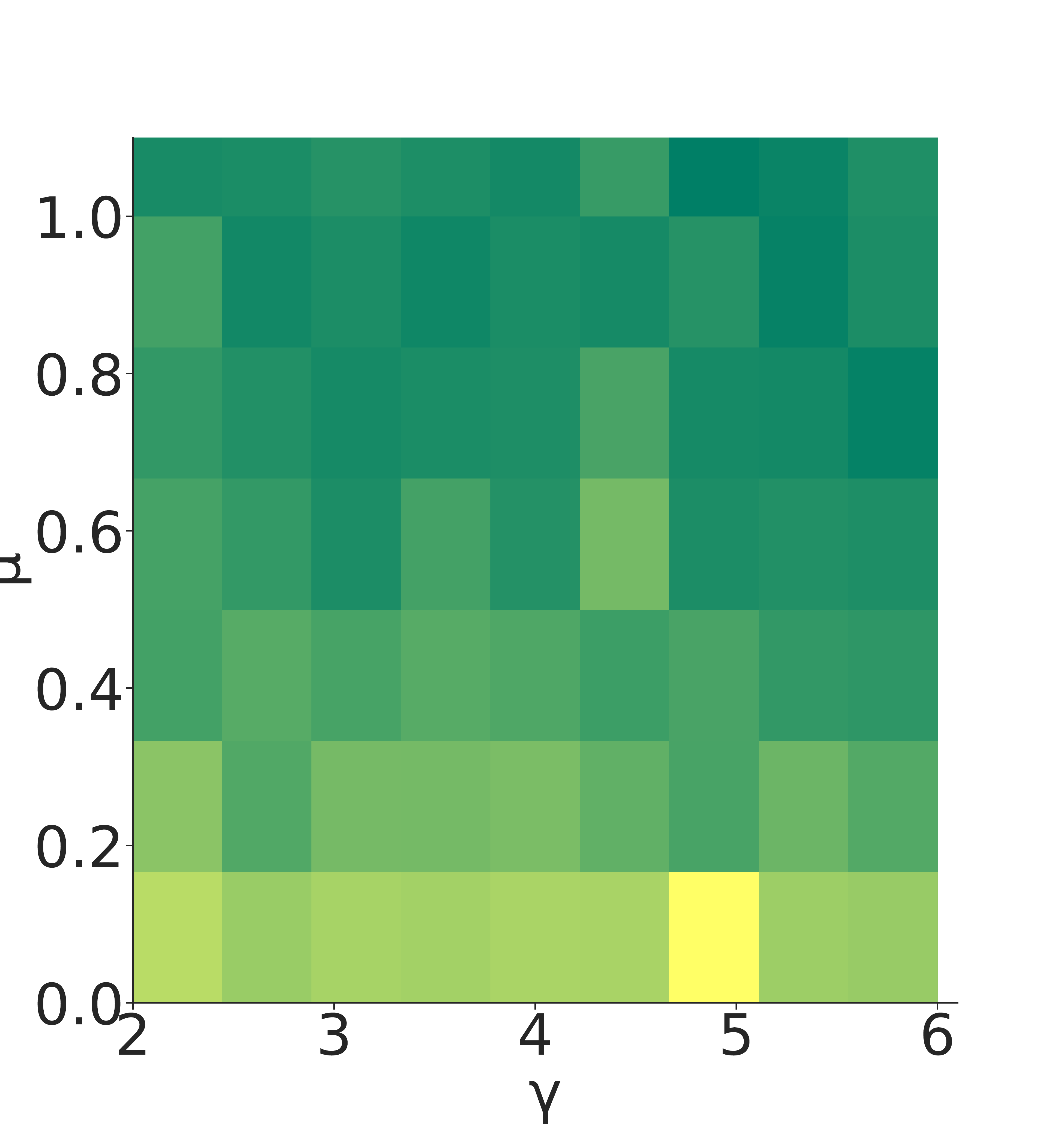} &
\includegraphics[width=0.25\textwidth,height=5cm]{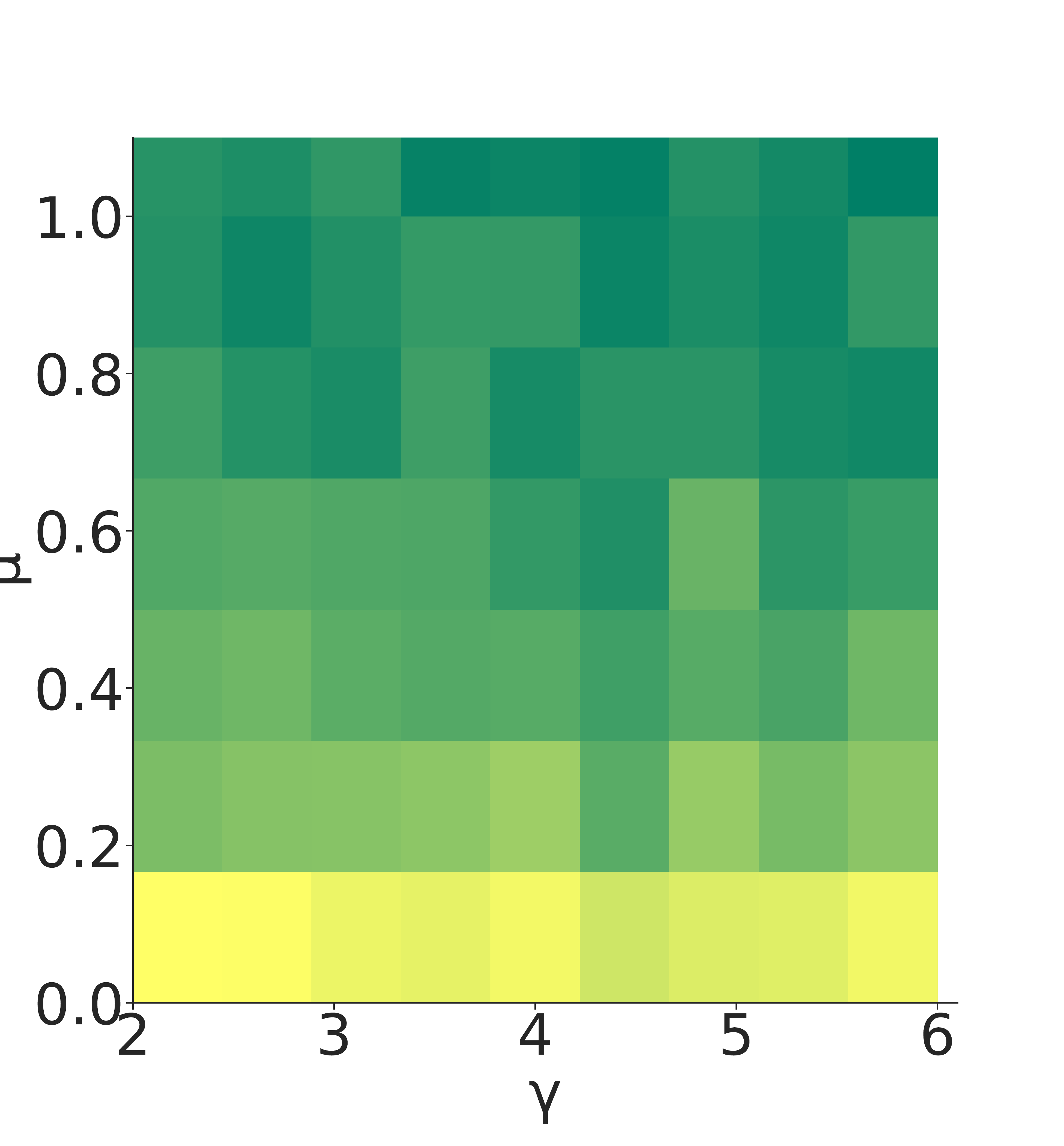} &
\includegraphics[width=0.25\textwidth,height=5cm]{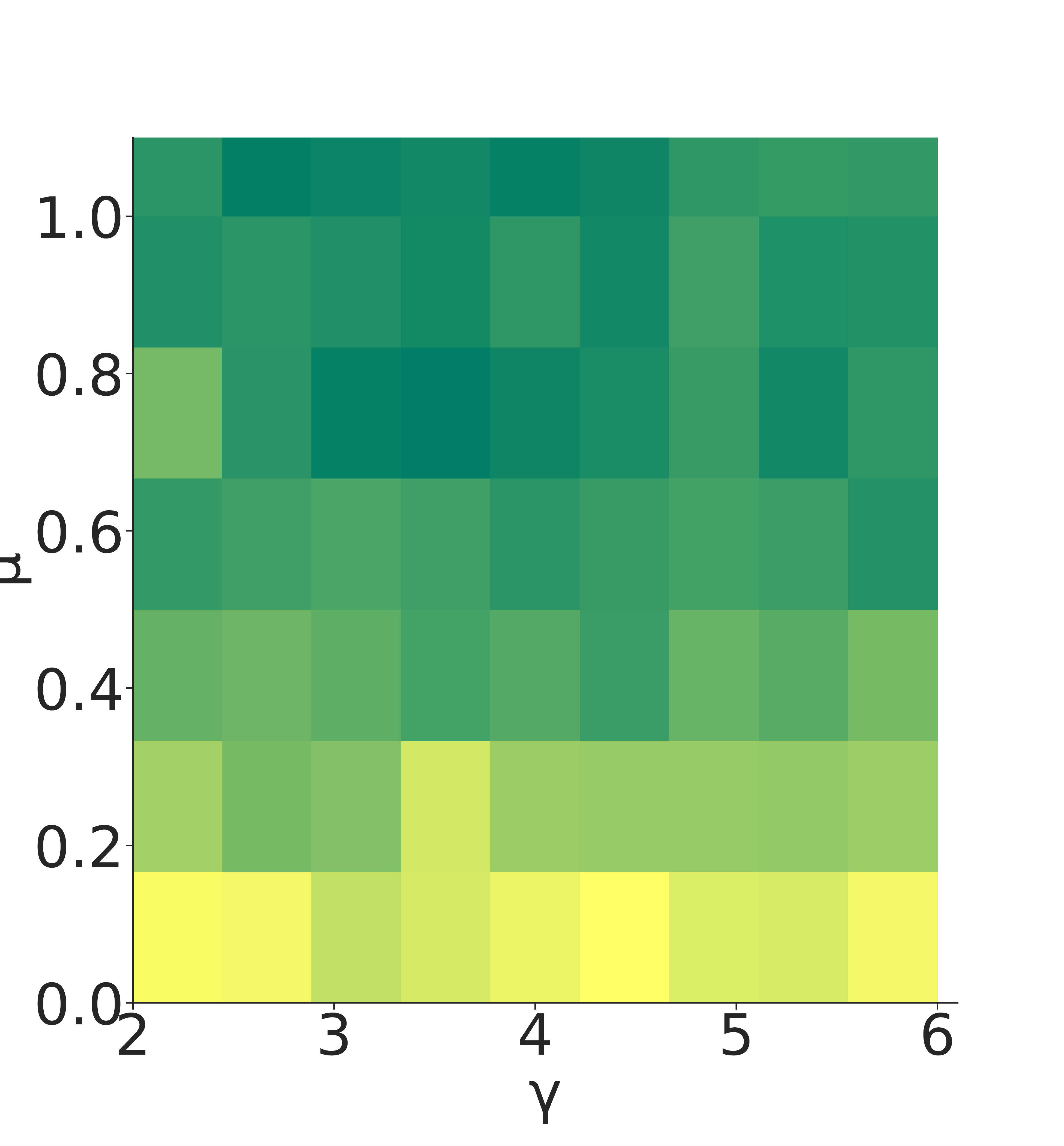} \\
\end{tabular}
\includegraphics[trim=19cm 12.1cm 23cm 13cm, clip=true, width=0.3\linewidth,height=2.2cm]{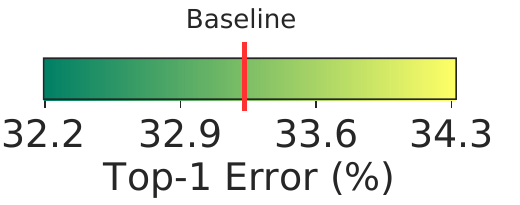}
\caption{Performance summary of static SFN configurations with $128$ nodes and various community numbers using a 5-layer MLP on the CIFAR-10 data set~\cite{cifar10}. We set the average degree $m = 3$ for the result reported here. (a)~A single community (the baseline result for the fully-connected network is indicated by the brown horizontal line) that is the recap of the result reported in Fig.~\ref{fig:SFN_single} for a horizontal stripe at $m = 3$, (b)~2 communities, (c)~3 communities, (d)~4 communities, (e)~5 communities, (f)~6 communities, (g)~7 communities, (h)~8 communities. In the cases of multiple communities (b)--(h), the parameters for the phase diagrams are the degree exponent $\gamma$ for each community~\cite{goh2001} and the mixing parameter $\mu$~\cite{lancichinetti2009community}, and the horizontal color bar with the baseline marked indicates the performance for those cases.}
\label{fig:SFN_results}
\end{figure*}

\begin{figure}
\includegraphics[width=\linewidth]{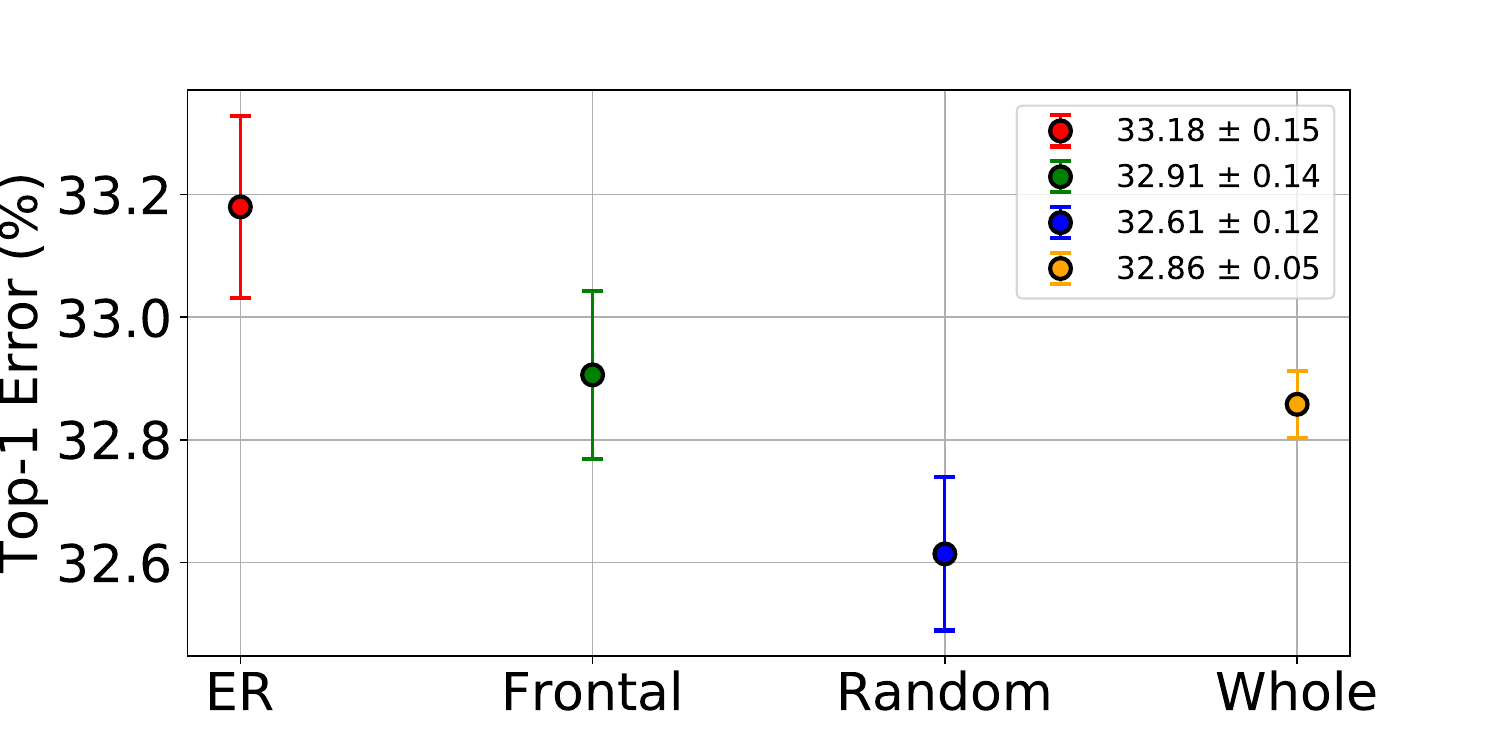}
\caption{Comparison of network performance of the \emph{C. elegans} network on the CIFAR-10 data set for three different structures: \emph{C. elegans} frontal neural network (``Frontal''), \emph{C. elegans} whole brain network (``Whole''), and a fixed random sample (``Random'') of 131 nodes drawn from the \emph{C. elegans} whole brain network, compared to a random network (``ER'') of similar size and connectivity. Error bars represent standard deviations computed across independent neural-network initialization seeds with the graph topology fixed.}
\label{fig:c_elegans}
\end{figure}

This section presents our findings, focusing on the top-1 error rates across different graph configurations and community structures. The top-1 error rate refers to the percentage of times the neural network's top prediction does not match the correct label (among the $10$ top-level categories of CIFAR-10~\cite{cifar10}). The results are averaged for each graph configuration, providing a detailed understanding of the relationship between network topology and neural network performance. Our comprehensive analysis reveals several important insights into network performance across different configurations. 

\subsection{Model networks without communities}
\label{sec:results_networks_without_communities}

In our study of model networks without communities, static scale-free networks~\cite{goh2001} exhibit a noteworthy enhancement in performance. The optimal relational graph, identified by the darkest color, achieves a 3\% improvement over the performance of the complete-graph baseline at the top-1 error $\approx 33.28\%$, with an average improvement of 1.3\% (Fig.~\ref{fig:SFN_single}). The existence of a ``sweet spot,'' marked by red rectangular regions with dashed borders, reveals that relational graphs with higher degree exponent $\gamma$ values---approaching those of ER random networks---and lower edge numbers with smaller values of $m$ (sparser networks) deliver superior performance. Similarly, the results from ER random networks~\cite{erdos1959random} exhibit a comparable trend, as shown in Fig.~\ref{fig:random_results}(a), where sparser networks consistently outperform their denser counterparts. This interpretation follows directly from statistical learning theory, where reducing model capacity relative to a fixed training set size improves generalization to unseen examples~\cite{vapnik1995nature}. This aligns with the broader understanding in network science that sparse connectivity can enhance the adaptability and generalization capability of neural systems by minimizing overfitting and improving signal propagation~\cite{Rusch2023ASO}.

\subsection{Model networks with communities}
\label{sec:results_networks_with_communities}

When examining model networks with communities~\cite{lancichinetti2009community}, we observe a significant impact on performance due to the introduction of community structures~\cite{Porter2009,Fortunato_review,Fortunato2022}. In ER random networks~\cite{erdos1959random} with communities, the implementation of any relational graph generally enhances the MLP architecture's predictive performance by 2.8\%, with the optimal graph outperforming the complete graph baseline by 5.3\% (Fig.~\ref{fig:random_results}). Importantly, when sweeping the mixing parameter $\mu$ across Figs.~\ref{fig:random_results}(b)--(h) and Figs.~\ref{fig:SFN_results}(b)--(h), the expected total number of edges---and therefore the total count of non-zero trainable parameters in the MLP---remains matched for a given base intra-community density ($p$ or $m$). Varying $\mu$ redistributes connectivity between intra- and inter-community channels without altering overall model capacity, confirming that the observed performance variations reflect the topological organization of weights rather than simple parameter reduction.

Notably, as the number of communities increases, the optimal graphs tend to shift towards configurations with greater inter-community density and edge density. This suggests that the presence of well-defined communities within a network imposes structured sparsity that regularizes learning and constrains the hypothesis space, reducing overfitting~\cite{Belkin2006ManifoldRA}.

For static networks~\cite{goh2001} with communities, relational graphs improve the MLP architecture's performance by 1.3\% on average, with the best graph surpassing the complete graph baseline by 3.3\% (we show the result for $m=3$ in Fig.~\ref{fig:SFN_results}, considering the range of ``sweet spot'' in Fig.~\ref{fig:SFN_single}). Similar to the ER random networks, the optimal graphs cluster in regions of higher inter-community density and degree exponents as the number of communities grows. A notable behavior observed is the strong correlation between network performance and mixing parameter $\mu$, rather than the degree distribution (Fig.~\ref{fig:correlation}). When $\mu \to 0$, communities are nearly isolated; while localized representations form, the absence of inter-community message passing prevents global coordination, yielding suboptimal task-level classification. As $\mu$ increases, the necessary inter-community routing is established to integrate modular representations globally, demonstrating that effective learning requires a balance between modular specialization and cross-module communication~\cite{kipf2016,Wu}.

\subsection{\emph{C. elegans} neural networks}
\label{sec:results_C_elegans}

For the biological neural network of \emph{C. elegans}, we observed superior performance compared to the frontal neural network (Fig.~\ref{fig:c_elegans}). This enhanced performance is likely attributable to the comprehensive connectivity and modular organization inherent in the \emph{C. elegans} brain network, which mirrors the efficiency observed in evolved biological systems. Interestingly, a fixed random subnetwork sample of 131 nodes drawn from the whole brain network also outperforms the frontal cortex network of the identical size. This serves as an existence demonstration that even an arbitrarily sampled, non-localized substructure from a well-organized biological connectome can retain significant functional capabilities, by constraining relevant information to be processed locally within modular neighborhoods before being integrated globally across them.

These results indicate that naturally evolved architectures possess inherent efficiency and robustness, offering valuable insights into the design of artificial neural networks. By emulating the modularity and connectivity patterns observed in biological networks, we can develop neural network architectures that are both efficient and adaptable, potentially leading to breakthroughs in machine learning applications. Overall, the insights gained from our analysis underscore the importance of network structure, particularly community and modular configurations, in enhancing neural network performance. These findings pave the way for future research focused on optimizing neural architectures through the strategic incorporation of community structures and connectivity patterns inspired by biological systems~\cite{sporns2000theoretical}.

\subsection{Effect of network depth}
\label{sec:network_depth}

\begin{figure}
    \begin{tabular}{l}
    (a) \\
    \includegraphics[trim=0cm 0cm 0cm 0cm, clip=true, width=0.65\columnwidth]{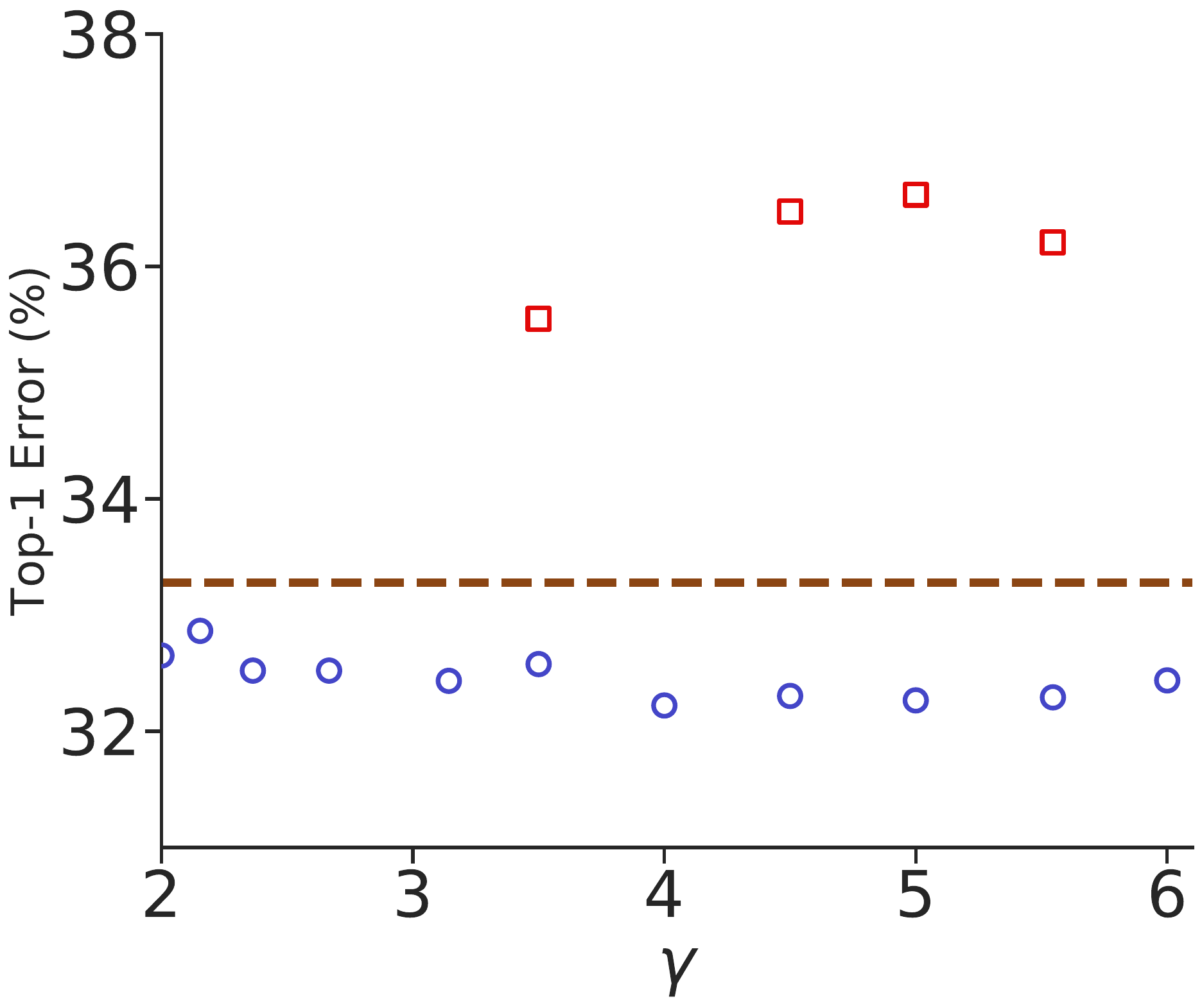} \\
    (b) \\
    \includegraphics[trim=0cm 0cm 0cm 0cm, clip=true, width=0.65\columnwidth]{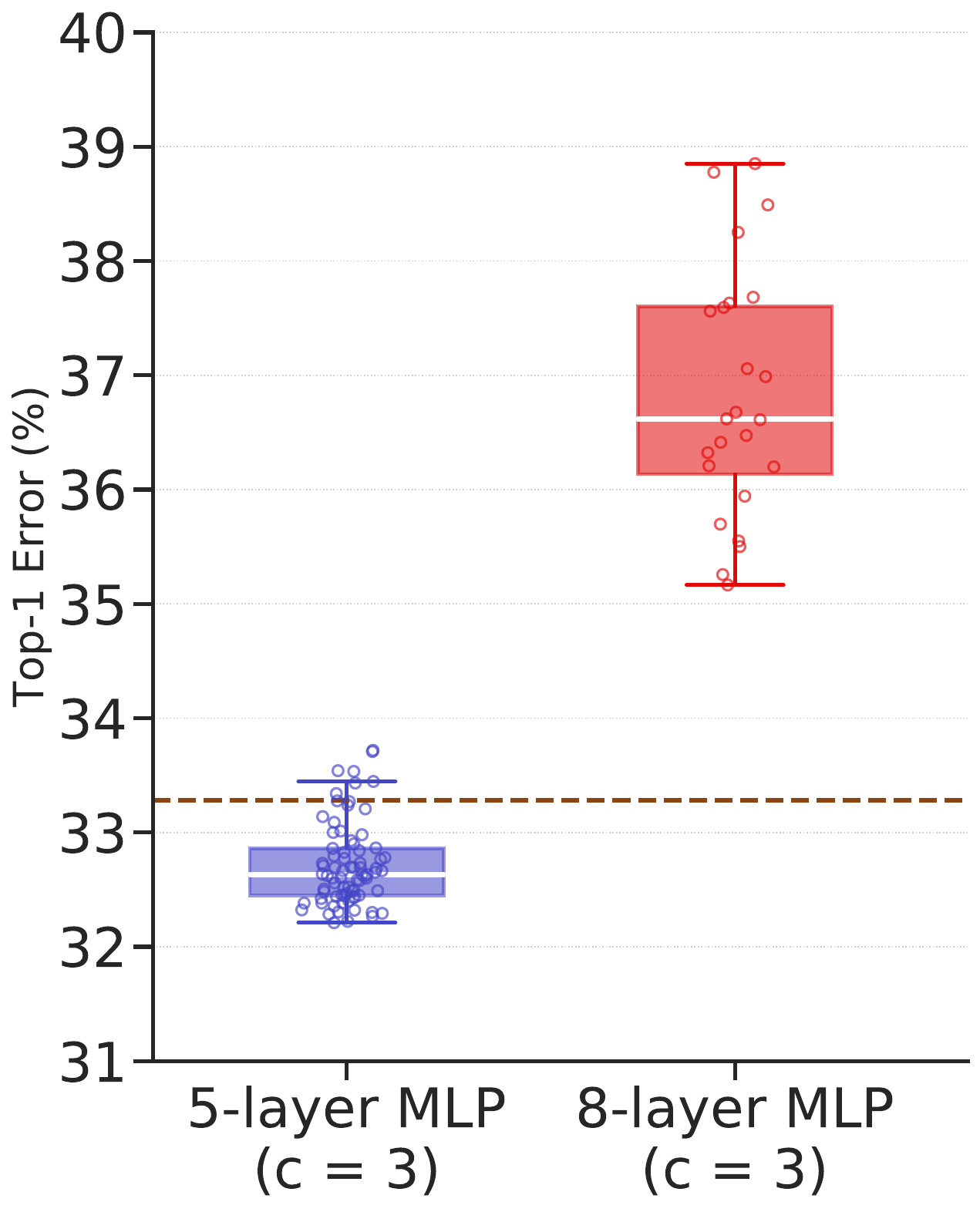}
    \end{tabular}
    \caption{Comparison of 5-layer and 8-layer fixed-width MLP performance for static SFN (with $m = 3$) configurations with community structures on the CIFAR-10 data set. (a)~Top-1 error percentage as a function of the degree exponent $\gamma$ at a fixed mixing parameter $\mu = 0.667$ and number of communities $c = 3$. Blue open circles denote the 5-layer MLP, while red open squares represent the 8-layer MLP. The horizontal brown dashed line marks the fully connected complete-graph baseline (top-1 error $\approx 33.28\%$). (b)~Paired Tukey boxplot distributions of top-1 errors across all tested $(\gamma, \mu)$ parameter combinations for the 5-layer versus 8-layer architectures at $c = 3$. The boxes span the interquartile range (IQR, 25th--75th percentiles), the horizontal white lines indicate the medians, and the whiskers extend to the most extreme data points within $1.5 \times \text{IQR}$ from the edge of the box. Individual run results are overlaid as small open circles to visualize variance.}
\label{fig:8-layer}
\end{figure}

To address the question of whether the findings reported in this section are specific to the 5-layer architecture or reflect more general structural principles, we extended the experimental design to include 8-layer MLPs for static scale-free networks both with and without communities. All simulations were conducted under identical conditions to the 5-layer experiments, with seeds increased from 5 to 10 to strengthen statistical robustness. The results reveal qualitatively new and theoretically significant behaviors. 

For static scale-free networks with community structures, increasing the depth from 5 to 8 layers produces a dramatic result: all tested configurations perform worse than the complete-graph baseline, as shown in Fig.~\ref{fig:8-layer}(a). This constitutes a categorical reversal of the 5-layer finding, where no configuration fell below baseline across the community structure parameter space. For Fig.~\ref{fig:8-layer}(b), we used a paired boxplot to visualize the distribution of top-1 errors across all tested $(\gamma,\mu)$ combinations for 5-layer versus 8-layer at $c = 3$. In addition to the categorical degradation in mean performance, this revealed that the 8-layer configurations exhibit substantially higher variance across the parameter space compared to 5-layer configurations, indicating that the relationship between topology and outcome is destabilized at greater depth.

We further note that Ref.~\cite{jiaxuan2020} explicitly identifies 8-layer architectures on CIFAR-10 as failure cases in their Appendix (``6. Discussion of Failure Cases''), reporting that the consistent sweet spot vanishes at this depth for CIFAR-10 but not ImageNet. Our 8-layer results independently reproduce this breakdown and extend it to community-structured networks, demonstrating that 5 layers represents an effective operational depth regime where modular benefits persist before deeper degradation sets in. 

While representation cosine similarity and feature shrinkage were not directly measured across hidden layers in this study, the observed degradation at 8 layers is consistent with the theoretical mechanics of over-smoothing in deep graph message passing, where successive rounds of neighborhood propagation cause representations to collapse toward indistinguishable states~\cite{li2018deeper,Oono2019GraphNN}. Compounding factors such as optimization bottlenecks and vanishing gradients common to un-residualized deep feedforward architectures may also contribute to this depth ceiling.

\subsection{Interplay between modularity and network depth}
\label{sec:results_summary}

\begin{figure*}
    \begin{tabular}{ll}
    \includegraphics[trim=0cm 1.4cm 0cm 0cm, clip=true, width=\textwidth]{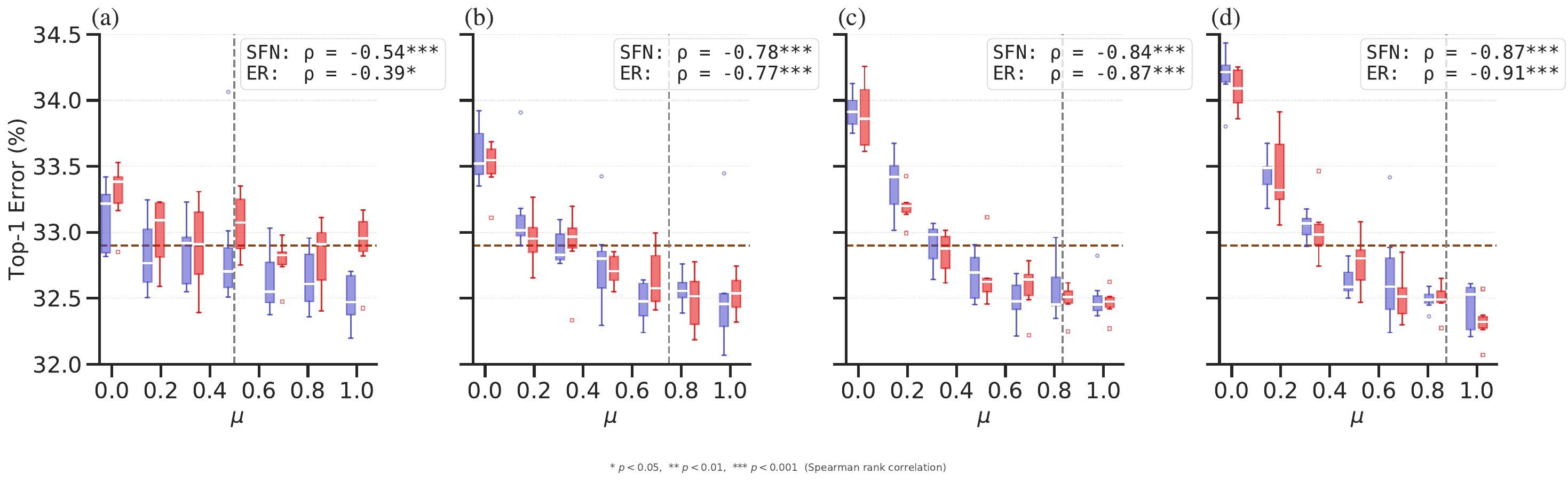}
    \end{tabular}
    \caption{Correlation between mixing parameter $\mu$ and network learning performance across different modular configurations. The boxplots summarize the distribution of top-1 classification error aggregated across the full suite of intra-community parameter values explored in previous simulations, with blue shaded boxes representing ER random networks (combining the complete set of tested internal connection probabilities $p \in [0.167,1.0]$) and red shaded boxes indicating static SFN with average degree $m = 3$ (combining the complete set of tested degree exponents $\gamma \in [2, 6]$). In each boxplot, the box spans the interquartile range (IQR, from the 25th percentile $Q_1$ to the 75th percentile $Q_3$), the central line denotes the median, whiskers extend to the most extreme data points located within $1.5 \times \text{IQR}$ of the box edges, and flier points indicate outliers lying beyond this range. Panels show results for variations with (a)~2 communities, (b)~4 communities, (c)~6 communities, and (d)~8 communities. The horizontal brown dashed line marks the fully connected complete-graph baseline, while the vertical gray dashed lines indicate the theoretical threshold $\mu = 1 - 1/c$ where the Newman--Girvan modularity approaches zero ($Q \to 0$), corresponding to a randomized null model without community preference. Spearman rank correlation coefficients $\rho$ between $\mu$ and the top-1 error are reported inside each inset text box for both network types ($^{*}p < 0.05$, $^{**}p < 0.01$, $^{***}p < 0.001$), confirming a statistically significant, negative monotonic relationship between inter-community density and classification error across all examined scenarios.}
\label{fig:correlation}
\end{figure*}

Overall, our findings underscore the pivotal role of community structures in enhancing neural network learning and performance within moderate-depth architectures. In our study of ER random networks and static SFNs, the best-performing relational graphs with community structures consistently outperformed the WS-flex generator~\cite{jiaxuan2020} across nearly all image classification tasks using a fixed-width 5-layer MLP on the CIFAR-10 data set. This superior performance can be understood within a rigorous statistical learning framework. Our results indicate that community structures provide a specific, higher-order form of structural regularization, yielding a kind of modular inductive bias. The presence of these communities imposes an architectural constraint where relevant informational features are processed locally within individual modules before being integrated globally across them, mirroring a hierarchical level of processing~\cite{ying2018,kipf2016,gilmer17a}.

A sparse network constrains which connections exist, reducing the number of learnable parameters, which limits model variance and outweighs the cost of the structural bias, thereby significantly improving out-of-sample generalization~\cite{tibshirani1996regression,kipf2016,han2015learning}. For the CIFAR-10 image classification task, this structural assumption is well-aligned with the intrinsic properties of the data. Image features are organized hierarchically into regional patterns (e.g., edges and textures), which must combine systematically into global object representations. A community-structured network, where message passing first operates densely within local subgraphs and then propagates globally across inter-community connections, enforces a computational hierarchy that matches this localized data structure. 

This functional mechanism is explicitly validated by the systematic empirical trends illustrated in Fig.~\ref{fig:correlation}. As shown by the clustered paired boxplots, increasing the mixing parameter $\mu$ from $0.0$ to $1.0$ triggers a continuous, monotonic reduction in top-1 classification error for both ER and scale-free configurations. To mathematically ground this observation, we report the non-parametric Spearman rank correlation coefficients $\rho$ within each panel. When the global architecture is partitioned into $c = 2$ communities [Fig.~\ref{fig:correlation}(a)], a significant negative correlation is already visible ($\rho_{\text{SFN}} = -0.54^{***}$ and $\rho_{\text{ER}} = -0.39^{*}$). Remarkably, as the mesoscale modularity becomes more granular, this topological dependence scales up drastically. For $c = 4$ [Fig.~\ref{fig:correlation}(b)] and $c = 6$ [Fig.~\ref{fig:correlation}(c)], the correlation sharpens ($\rho_{\text{SFN}} = -0.78^{***}$ and $-0.84^{***}$, respectively), with the ER network converging to match this structural sensitivity ($\rho_{\text{ER}} = -0.87^{***}$). At our highest tested granularity of $c = 8$ communities [Fig.~\ref{fig:correlation}(d)], the monotonic coupling reaches its maximum intensity ($\rho_{\text{SFN}} = -0.87^{***}$ and $\rho_{\text{ER}} = -0.91^{***}$), indicating that nearly $91\%$ of the variance in the classification error distribution is bounded by the macro-level routing parameter $\mu$. This systematic behavior demonstrates that robust inter-community connectivity is essential for integrating locally processed representations into a globally coherent classification.

Intriguingly, when viewed through the lens of formal modularity $Q \approx (1 - \mu) - 1/c$, this empirical optimum at high $\mu$ suggests that the best-performing relational topologies reside in regions of relatively modest or small modularity, near or beyond the randomized null threshold $\mu = 1 - 1/c$ [marked by the vertical dashed lines in Fig.~\ref{fig:correlation}], where $Q \to 0$ or even becomes negative ($\mu > 1 - 1/c$). While our result does not imply that completely unstructured networks are unconditionally optimal, it indicates that only a mild degree of modular segregation is required to regularize local representations, whereas preserving strong global integration paths across modules is critical to avoiding informational bottlenecks in classification. On the contrary, a traditional fully connected network with $N$ nodes per layer possesses $N^2$ learnable weights per layer. While this provides the network with maximal representation capacity, it introduces maximal variance, leading to rapid overfitting and poor generalization on unseen data.

An instructive boundary condition emerges when exploring deep 8-layer MLPs, where the structural benefits of these community configurations undergo a complete, depth-dependent reversal. We interpret this phenomenon through the lens of graph message-passing mechanics: each successive layer of message exchange expands a node's structural neighborhood. After $r$ hidden layers, a node's representation integrates information from all nodes situated within a graph geodesic distance $r$. As $r$ increases to deep thresholds, the individual receptive fields heavily overlap, which theoretically drives hidden representations toward uniform vectors across the graph volume. This represents a plausible explanation for the performance collapse observed at 8 layers, where all configurations uniformly degrade well below the baseline complete-graph reference line~\cite{li2018deeper,Oono2019GraphNN}. Consequently, moderate depths such as 5 layers operate within a viable regime that balances feature depth against over-smoothing and optimization degradation.

In conclusion, this detailed phase-space mapping demonstrates that network performance is tightly coupled to the mixing parameter $\mu$, confirming that modular partitioning facilitates robust learning when balanced by sufficient cross-community communication (Fig.~\ref{fig:correlation}). This finding is further supported by our baseline comparisons with the naturally optimized \emph{C. elegans} neural network~\cite{bertolero2015modular}, underscoring that the mesoscale wiring diagram of an underlying graph acts as an important determinant of deep learning efficiency and confirming that naturally evolved architectures encode structural priors favorable to predictive generalization~\cite{dressler2010survey}.

\section{Conclusion and Discussions}
\label{sec:conclusion}

In this study, we have investigated the relationship between the underlying graph structure of neural networks and their predictive performance under a statistical learning framework. By employing the relational graph methodology and extending previous work with WS-flex generators~\cite{jiaxuan2020} to encompass more complex and realistic networks, we have systematically incorporated community structures within relational graphs. Our findings reveal a direct correlation between the inter- and intra-community connectivity and network performance. Moreover, introducing communities into sparse subnetworks can match or exceed the performance of their fully-connected and non-community counterparts specifically through structural regularization~\cite{frankle2018the}. Within moderate-depth architectures (such as 5-layer MLPs on CIFAR-10), sparse structured relational graphs consistently match or outperform the dense fully connected baseline.

This advantage can be attributed to several interrelated architectural factors: sparser structured configurations outperform the dense fully connected baseline at moderate depths, confirming that unconstrained parameter capacity often degrades out-of-sample generalization. Furthermore, embedding explicit community divisions yields a mesoscale inductive bias that surpasses prior relational graph benchmarks, such as the standard small-world WS-flex generator, while topological modularity constraints implicitly narrow the network's hypothesis space to mitigate overfitting without relying on parameter norm penalties. This structural dependence provides interpretable design principles, highlighting the mixing parameter $\mu$ as a primary architectural lever that must maintain sufficient cross-community connectivity for global feature integration. Crucially, these modular structural priors---naturally inspired by evolved biological networks such as \emph{C.~elegans}---translate effectively into artificial neural architectures, though our depth exploration demonstrates that these advantages are strictly bounded by network depth, undergoing a performance reversal at 8 layers consistent with over-smoothing and optimization constraints.

These findings align with biological systems, where modularity is naturally encoded to support specialized processing and robust integration. Community-aware graph design presents a promising direction for neural architecture search and the design of efficient graph neural networks (GNNs)~\cite{you2020design,dressler2010survey,liu2024community}. Future work should investigate how these modular design principles generalize to other data sets, deeper residual architectures, and dynamic or automated community detection methods during training.

\begin{acknowledgments}
The authors sincerely appreciate the authors of Ref.~\cite{jiaxuan2020} for sharing their codes online~\cite{graph2nn}. S.H.L. is grateful to Dongseong Hwang and Jae Oh Woo for their opinions and discussions in regard to this topic at the conceiving stage of the work, and acknowledges the efforts by Seora Son and Eun Ji Choi with their initial assessment of \texttt{graph2nn}. This work was supported by the National Research Foundation (NRF) of Korea under Grant No.~RS-2026-25468383, and results of a study on the ``Gyeongsangnam-do Regional Innovation System \& Education (RISE)'' Project, supported by the Ministry of Education and Gyeongsangnam-do.
\end{acknowledgments}


\bibliography{references}

@PREAMBLE{
 "\providecommand{\noopsort}[1]{}" 
 # "\providecommand{\singleletter}[1]{#1}%" 
}

@article{rumelhart1986,
  title={Learning representations by back-propagating errors},
  author={Rumelhart, D. E. and Hinton, G. E. and Williams, R. J.},
  journal={Nature},
  volume={323},
  number={6088},
  pages={533-536},
  year={1986}
}

@inproceedings{kipf2016,
title={Semi-Supervised Classification with Graph Convolutional Networks},
author={Thomas N. Kipf and Max Welling},
booktitle={International Conference on Learning Representations},
year={2017},
url={https://openreview.net/forum?id=SJU4ayYgl}
}

@inproceedings{jiaxuan2020, author = {You, Jiaxuan and Leskovec, Jure and He, Kaiming and Xie, Saining}, title = {Graph structure of neural networks}, year = {2020}, publisher = {JMLR.org}, booktitle = {Proceedings of the 37th International Conference on Machine Learning}, articleno = {1009}, numpages = {11}, series = {ICML'20} }

@article{Wu,
  title={A comprehensive survey on graph neural networks},
  author={Wu, Z. and Pan, S. and Chen, F. and Long, G. and Zhang, C. and Yu, P. S.},
  journal={IEEE transactions on neural networks and learning systems},
  volume={32},
  number={1},
  pages={4-24},
  year={2020}
}

@article{goh2001,
  title={Universal behavior of load distribution in scale-free networks},
  author={Goh, K.-I. and Kahng, B. and Kim, D.},
  journal={Phys. Rev. Lett.},
  volume={87},
  pages={270701},
  year={2001}
}

@article{mehta2019,
  title={A high-bias, low-variance introduction to machine learning for physicists},
  author={Mehta, P. and Bukov, M. and Wang, C. H. and Day, A. G. and Richardson, C. and Fisher, C. K. and Schwab, D. J.},
  journal={Physics Reports},
  volume={810},
  pages={1-124},
  year={2019}
}

@inproceedings{ying2018,
  author    = {Ying, Rex and You, Jiaxuan and Morris, Christopher and Ren, Xiang and Hamilton, William L. and Leskovec, Jure},
  title     = {Hierarchical graph representation learning with differentiable pooling},
  booktitle = {Proceedings of the 32nd International Conference on Neural Information Processing Systems},
  series    = {NIPS'18},
  pages     = {4805--4815},
  year      = {2018},
  publisher = {Curran Associates Inc.},
  address   = {Red Hook, NY, USA}
}

@inproceedings{xie2019exploring,
  title={Exploring randomly wired neural networks for image recognition},
  author={Xie, Saining and Kirillov, Alexander and Girshick, Ross and He, Kaiming},
  booktitle={Proceedings of the IEEE/CVF International Conference on Computer Vision},
  pages={1284--1293},
  year={2019}
}

@article{lancichinetti2009community,
  title = {Community detection algorithms: A comparative analysis},
  author = {Lancichinetti, Andrea and Fortunato, Santo},
  journal = {Phys. Rev. E},
  volume = {80},
  issue = {5},
  pages = {056117},
  numpages = {11},
  year = {2009},
  month = {Nov},
  publisher = {American Physical Society},
  doi = {10.1103/PhysRevE.80.056117},
  url = {https://link.aps.org/doi/10.1103/PhysRevE.80.056117}
}

@article{bertolero2015modular,
  title={The modular and integrative functional architecture of the human brain},
  author={Bertolero, Maxwell A and Yeo, BT Thomas and D'Esposito, Mark},
  journal={Proceedings of the National Academy of Sciences},
  volume={112},
  number={49},
  pages={E6798--E6807},
  year={2015},
  publisher={National Academy of Sciences}
}

@article{brenner1974genetics,
  title={The genetics of Caenorhabditis elegans},
  author={Brenner, Sydney},
  journal={Genetics},
  volume={77},
  number={1},
  pages={71--94},
  year={1974},
  publisher={Oxford University Press}
}

@InProceedings{gilmer17a,
  title={Neural Message Passing for Quantum Chemistry},
  author={Justin Gilmer and Samuel S. Schoenholz and Patrick F. Riley and Oriol Vinyals and George E. Dahl},
  booktitle={Proceedings of the 34th International Conference on Machine Learning},
  pages={1263--1272},
  year={2017},
  editor={Precup, Doina and Teh, Yee Whye},
  volume={70},
  series={Proceedings of Machine Learning Research},
  month={06--11 Aug},
  publisher={PMLR},
}

@article{you2020design,
  title={Design space for graph neural networks},
  author={You, Jiaxuan and Ying, Zhitao and Leskovec, Jure},
  journal={Advances in Neural Information Processing Systems},
  volume={33},
  pages={17009--17021},
  year={2020}
}

@article{kaiser2006nonoptimal,
  title={Nonoptimal component placement, but short processing paths, due to long-distance projections in neural systems},
  author={Kaiser, Marcus and Hilgetag, Claus C},
  journal={PLoS Comput. Biol.},
  volume={2},
  number={7},
  pages={e95},
  year={2006},
  publisher={Public Library of Science San Francisco, USA}
}

@article{liu2024community,
  title={A community detection and graph-neural-network-based link prediction approach for scientific literature},
  author={Liu, Chunjiang and Han, Yikun and Xu, Haiyun and Yang, Shihan and Wang, Kaidi and Su, Yongye},
  journal={Mathematics},
  volume={12},
  number={3},
  pages={369},
  year={2024},
  publisher={MDPI}
}

@article{dressler2010survey,
  title={A survey on bio-inspired networking},
  author={Dressler, Falko and Akan, Ozgur B},
  journal={Computer networks},
  volume={54},
  number={6},
  pages={881--900},
  year={2010},
  publisher={Elsevier}
}

@misc{YA_relational_graph,
  author = {Arya, Yash},
  title = {{relational\_graph\_web}},
  howpublished = "\url{https://github.com/yasharyaa/relational_graph_web.git}",
  year = 2025,
  note = {Accessed: 2025-07-12}
}

@misc{YA_LFR,
  author = {Arya, Yash},
  title = {{Simplified\_LFR\_Benchmark\_Graph}},
  howpublished = "\url{https://github.com/yasharyaa/Simplified_LFR_Benchmark_Graph.git}",
  year = 2025,
  note = {Accessed: 2025-07-12}
}

@techreport{cifar10,
  title={Learning Multiple Layers of Features from Tiny Images},
  author={Krizhevsky, Alex},
  year={2009},
  institution={University of Toronto},
  address={Toronto, Canada},
  url={https://www.cs.toronto.edu/~kriz/learning-features-2009-TR.pdf}
}

@article{MLP,
  author = {Rosenblatt, Frank},
  title = {The Perceptron: A Probabilistic Model for Information Storage and Organization in the Brain},
  journal = {Psychological Review},
  year = {1958},
  volume = {65},
  number = {6},
  pages = {386-408},
  doi = {10.1037/h0042519}
}

@article{CNN,
  author = {Fukushima, Kunihiko},
  title = {Neocognitron: A Self-organizing Neural Network Model for a Mechanism of Pattern Recognition Unaffected by Shift in Position},
  journal = {Biological Cybernetics},
  year = {1980},
  volume = {36},
  pages = {193-202},
  doi = {10.1007/BF00344251}
}

@inproceedings{ResNet,
  title={Deep Residual Learning for Image Recognition},
  author={He, Kaiming and Zhang, Xiangyu and Ren, Shaoqing and Sun, Jian},
  booktitle={Proceedings of the IEEE Conference on Computer Vision and Pattern Recognition (CVPR)},
  year={2016},
  pages={770-778},
  doi={10.1109/CVPR.2016.90}
}

@inproceedings{ImageNet,
  title={ImageNet: A Large-Scale Hierarchical Image Database},
  author={Deng, Jia and Dong, Wei and Socher, Richard and Li, Li-Jia and Li, Kai and Fei-Fei, Li},
  booktitle={2009 IEEE Conference on Computer Vision and Pattern Recognition},
  pages={248-255},
  year={2009},
  organization={IEEE},
  doi={10.1109/CVPR.2009.5206848}
}

@article{watts1998collective,
  title={Collective dynamics of `small-world' networks},
  author={Watts, Duncan J. and Strogatz, Steven H.},
  journal={Nature},
  volume={393},
  number={6684},
  pages={440--442},
  year={1998},
  publisher={Nature Publishing Group},
  doi={10.1038/30918}
}

@book{NewmanBook,
	Address = {Oxford, United Kingdom},
	Author = {Newman, M. E. J.},
	Isbn = {978-0198805090},
	Publisher = {Oxford University Press},
	Title = {Networks},
	Year = {2018}}

@BOOK{FirstCourseBook,
  title     = "A first course in network science",
  author    = "Menczer, Filippo and Fortunato, Santo and Davis, Clayton A",
  publisher = "Cambridge University Press",
  month     =  jan,
  year      =  2020,
  address   = "Cambridge, England"
}

@book{BarabasiBook,
  title={Network Science},
  author={Barab{\'a}si, Albert-L{\'a}szl{\'o}},
  year={2016},
  publisher={Cambridge University Press},
  address={Cambridge, UK},
  isbn={9781107076266}
}

@misc{graph2nn,
  author = {{Meta (formerly Facebook) Research}},
  title = {{Code for the paper ``Graph Structure of Neural Networks''}},
  howpublished = {\url{https://github.com/facebookresearch/graph2nn}},
  note = {Accessed: 2025-07-12}
}

@article{barabasi1999emergence,
  title={Emergence of scaling in random networks},
  author={Barab{\'a}si, Albert-L{\'a}szl{\'o} and Albert, R{\'e}ka},
  journal={Science},
  volume={286},
  number={5439},
  pages={509--512},
  year={1999},
  publisher={American Association for the Advancement of Science},
  doi={10.1126/science.286.5439.509}
}

@article{erdos1959random,
  author = {Erd\H{o}s, Paul and R\'enyi, Alfr\'ed},
  title = {On Random Graphs {I}},
  journal = {Publicationes Mathematicae},
  volume = {6},
  pages = {290--297},
  year = {1959}
}

@article{Porter2009,
author = {Porter, Mason A. and Onnela, Jukka-Pekka and Mucha, Peter J.},
title = {Communities in Networks},
journal = {Not. Am. Math. Soc.},
year = {2009},
volume = {56},
number = {9},
pages = {1082-1097},
url = {https://www.ams.org/journals/notices/200909/rtx090901082p.pdf}
}

@article{Fortunato2022,
  author = {Fortunato, Santo and Newman, Mark E. J.},
  year = {2022},
  date = {2022/08/01},
  title = {20 years of network community detection},
  journal = {Nature Physics},
  pages = {848--850},
  volume = {18},
  number = {8},
  issn = {1745-2481},
  url = {https://doi.org/10.1038/s41567-022-01716-7},
  doi = {10.1038/s41567-022-01716-7},
}

@article{Fortunato_review,
title = {Community detection in graphs},
journal = {Phys. Rep.},
volume = {486},
number = {3},
pages = {75--174},
year = {2010},
issn = {0370-1573},
doi = {https://doi.org/10.1016/j.physrep.2009.11.002},
author = {Santo Fortunato}
}

@article{bullmore2009complex,
  title={Complex brain networks: graph theoretical analysis of structural and functional systems},
  author={Bullmore, Ed and Sporns, Olaf},
  journal={Nature reviews neuroscience},
  volume={10},
  number={3},
  pages={186--198},
  year={2009},
  publisher={Nature Publishing Group UK London}
}

@article{meunier2010modular,
  title={Modular and hierarchically modular organization of brain networks},
  author={Meunier, David and Lambiotte, Renaud and Bullmore, Edward T},
  journal={Frontiers in neuroscience},
  volume={4},
  pages={200},
  year={2010},
  publisher={Frontiers Research Foundation}
}

@article{clune2013evolutionary,
  title={The evolutionary origins of modularity},
  author={Clune, Jeff and Mouret, Jean-Baptiste and Lipson, Hod},
  journal={Proceedings of the Royal Society b: Biological sciences},
  volume={280},
  number={1755},
  pages={20122863},
  year={2013}
}

@inproceedings{
frankle2018the,
title={The Lottery Ticket Hypothesis: Finding Sparse, Trainable Neural Networks},
author={Jonathan Frankle and Michael Carbin},
booktitle={International Conference on Learning Representations},
year={2019},
url={https://openreview.net/forum?id=rJl-b3RcF7},
}

@inproceedings{li2018deeper,
  title={Deeper insights into graph convolutional networks for semi-supervised learning},
  author={Li, Qimai and Han, Zhichao and Wu, Xiao-Ming},
  booktitle={Proceedings of the AAAI conference on artificial intelligence},
  volume={32},
  number={1},
  year={2018}
}

@article{Oono2019GraphNN,
  title={Graph Neural Networks Exponentially Lose Expressive Power for Node Classification},
  author={Kenta Oono and Taiji Suzuki},
  journal={arXiv:1905.10947},
  year={2019},
  url={https://api.semanticscholar.org/CorpusID:209994765}
}

@article{sporns2000theoretical,
  author     = {Sporns, Olaf and Tononi, Giulio and Edelman, Gerald M.},
  title      = {Theoretical Neuroanatomy: Relating Anatomical and Functional Connectivity in Graphs and Cortical Connection Matrices},
  journal    = {Cerebral Cortex},
  volume     = {10},
  number     = {2},
  pages      = {127--141},
  year       = {2000},
  month      = {02},
  doi        = {10.1093/cercor/10.2.127},
  url        = {https://doi.org/10.1093/cercor/10.2.127}
}

@article{Rusch2023ASO,
  title={A Survey on Oversmoothing in Graph Neural Networks},
  author={T. Konstantin Rusch and Michael M. Bronstein and Siddhartha Mishra},
  journal={ArXiv},
  year={2023},
  volume={abs/2303.10993},
  url={https://api.semanticscholar.org/CorpusID:257632346}
}

@article{LeCun2015DeepLearning,
  author    = {Yann LeCun and Yoshua Bengio and Geoffrey Hinton},
  title     = {Deep Learning},
  journal   = {Nature},
  volume     = {521},
  number     = {7553},
  pages      = {436--444},
  year       = {2015},
  publisher  = {Nature Publishing Group},
  doi        = {10.1038/nature14539}
}

@inproceedings{zoph2017neural,
  title={Neural Architecture Search with Reinforcement Learning},
  author={Zoph, Barret and Le, Quoc V.},
  booktitle={International Conference on Learning Representations (ICLR)},
  year={2017}
}

@book{vapnik1995nature,
  title={The Nature of Statistical Learning Theory},
  author={Vapnik, Vladimir N.},
  year={1995},
  publisher={Springer},
  address={New York},
  doi={10.1007/978-1-4757-2440-0}
}

@article{Belkin2006ManifoldRA,
  title={Manifold Regularization: A Geometric Framework for Learning from Labeled and Unlabeled Examples},
  author={Mikhail Belkin and Partha Niyogi and Vikas Sindhwani},
  journal={J. Mach. Learn. Res.},
  year={2006},
  volume={7},
  pages={2399-2434},
  url={https://api.semanticscholar.org/CorpusID:16902615}
}

@article{tibshirani1996regression,
  title={Regression Shrinkage and Selection via the Lasso},
  author={Tibshirani, Robert},
  journal={Journal of the Royal Statistical Society: Series B},
  volume={58},
  number={1},
  pages={267--288},
  year={1996},
  publisher={Wiley}
}

@article{han2015learning,
  title={Learning both Weights and Connections for Efficient Neural Networks},
  author={Han, Song and Pool, Jeff and Tran, John and Dally, William},
  journal={Advances in Neural Information Processing Systems},
  volume={28},
  year={2015}
}

\end{document}